\documentclass{article}

\usepackage[preprint]{neurips_2024}

\usepackage[numbers]{natbib}
\usepackage[dvipsnames]{xcolor}
\usepackage[colorlinks]{hyperref}
\hypersetup{
linkcolor=BrickRed
,citecolor=NavyBlue
,filecolor=Mulberry
,urlcolor=NavyBlue
,menucolor=BrickRed
,runcolor=Mulberry
,linkbordercolor=BrickRed
,citebordercolor=NavyBlue
,filebordercolor=Mulberry
,urlbordercolor=NavyBlue
,menubordercolor=BrickRed
,runbordercolor=Mulberry
}

\usepackage[utf8]{inputenc} 
\usepackage[T1]{fontenc}    
\usepackage{hyperref}       
\usepackage{url}            
\usepackage{booktabs}       
\usepackage{amsfonts}       
\usepackage{nicefrac}       
\usepackage{microtype}      
\usepackage{xcolor}         
\usepackage{xspace}
\usepackage{amsmath}
\usepackage{graphicx}
\usepackage{subcaption}
\usepackage{caption}
\usepackage{array}
\usepackage{CJKutf8}
\usepackage{multirow}
\usepackage{booktabs}
\usepackage{verbatim}
\usepackage{multicol}
\usepackage{colortbl}
\usepackage{float}
\usepackage{enumitem}

\newcommand{\method}{CLASI\xspace}
\newcommand{\benchmark}{RealSI\xspace}
\newcommand{\humaneval}{VIP\xspace}
\newcommand{\product}{Commerical 4\xspace}
\newcommand{\website}{https://byteresearchcla.github.io/clasi}

\usepackage{xcolor}
\usepackage{booktabs,nicematrix}
\usepackage{colortbl} 
\usepackage{booktabs,xcolor,siunitx}

\usepackage{tikz,lipsum,lmodern}
\usepackage[most]{tcolorbox}
\usepackage{tcolorbox}
\newtcolorbox{mybox}{colframe = red!75!black}

\usepackage{graphicx} 
\usepackage{fancyhdr} 
\usepackage{longtable}

\usepackage{amsmath,amsfonts,bm, amssymb}
\usepackage{cleveref}
\usepackage{eucal}

\def\eqref#1{equation~\ref{#1}}

\def\1{\bm{1}}

\def\rvI{{\mathbf{I}}}

\def\rvk{{\mathbf{k}}}

\def\rvs{{\mathbf{s}}}

\def\rvx{{\mathbf{x}}}
\def\rvy{{\mathbf{y}}}

\title{
Towards Achieving Human Parity on  End-to-end Simultaneous  Speech Translation via LLM Agent
}

\author{
Cross Language Agent Team \\
\vskip 0.05in
ByteDance Research
}

\usepackage{fancyhdr}
\fancyheadoffset[loh,reh]{20mm}
\begin{document}
\maketitle
\thispagestyle{fancy}

\begin{CJK}{UTF8}{gbsn}
\begin{abstract}
  In this paper, we present \textbf{C}ross \textbf{L}anguage \textbf{A}gent - \textbf{S}imultaneous \textbf{I}nterpretation, \method, a high-quality and human-like Simultaneous Speech Translation (SiST)\footnote{In this paper, we use Simultaneous Interpretation and Simultaneous Speech Translation interchangeably.} System.
  Inspired by professional human interpreters, we utilize a novel data-driven read-write strategy to balance the translation quality and latency.
  To address the challenge of translating in-domain terminologies, \method employs a multi-modal retrieving module to obtain relevant information to augment the translation.
  Supported by LLMs, our approach can generate error-tolerated translation by considering the input audio, historical context, and retrieved information.
  Experimental results show that our system outperforms other systems by significant margins. Aligned with professional human interpreters, we evaluate {\method} with a better human evaluation metric, valid information proportion~(\humaneval), which measures the amount of information that can be successfully conveyed to the listeners. 
  In the real-world scenarios, where the speeches are often disfluent, informal, and unclear, \method achieves \humaneval of 81.3\% and 78.0\% for Chinese-to-English and English-to-Chinese translation directions, respectively. In contrast, state-of-the-art commercial or open-source systems only achieve 35.4\% and 41.6\%. On the extremely hard dataset, where other systems achieve under 13\% \humaneval, \method can still achieve 70\% \humaneval. Demonstrations and human-annotated test sets are available at \url{\website}.
\end{abstract}

\begin{figure}[h]
    \centering
    \includegraphics[width=1.0\linewidth]{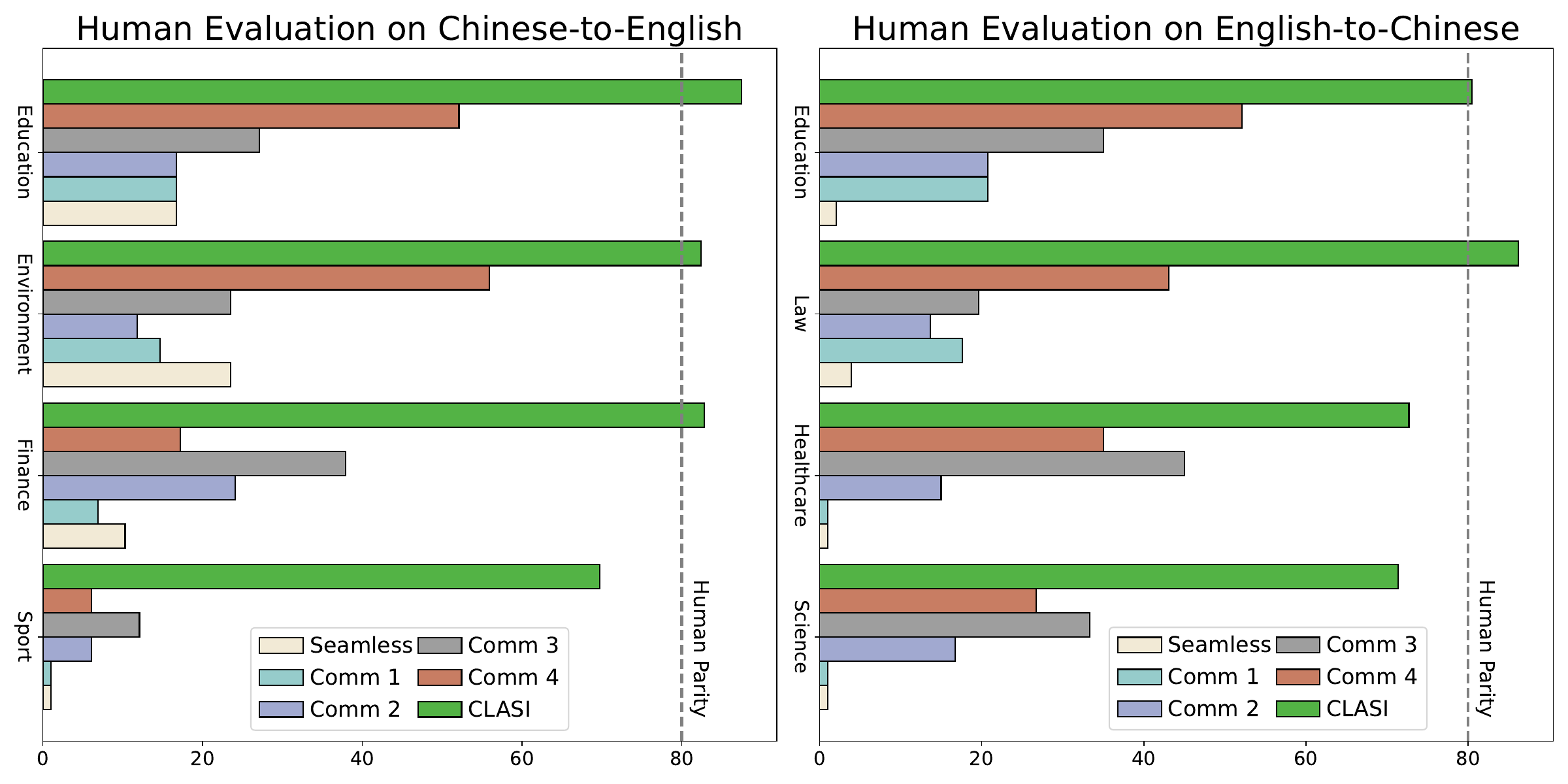}
    \caption{Performance evaluation. \method significantly outperforms the leading commercial and open-source systems using a more reliable \humaneval metric, achieving human interpreter parity. }
    \label{fig:fig1}
\end{figure}

\section{Introduction}

Simultaneous speech translation (SiST) is recognized as one of the most challenging tasks in the translation domain \cite{jones2014conference}. 
Machine-assisted automatic interpretation has been receiving much attention in the natural language processing (NLP) community \cite{iwslt-2021-international, iwslt-2020-international, iwslt-2022-international, iwslt-2023-international}. 
Traditional simultaneous translation approaches \cite{cho2016can,gu2017learning,zhao2021volctrans} 
usually employs a cascaded system, involving a streaming Automatic Speech Recognition (ASR) model, a punctuation model and a Machine Translation (MT) model.
However, such cascaded systems often suffer error propagation and latency from the ASR module. 
Despite these advancements in both academic SiST models \cite{barrault2023seamless, fukuda-etal-2023-naist, liu2024recentadvancesendtoendsimultaneous, papi-etal-2023-direct, ren2020simulspeech, zeng-etal-2021-realtrans, zhang2023end} and commercial SiST engines, the translation quality is still far from satisfactory. 
As shown in \Cref{fig:fig1}, we conduct a human assessment of the current accessible SiST systems. From the user-centered perspective, these systems only deliver less than 42\% of the valid information to listeners, which heavily affects communication effectiveness. 
In contrast, professional human interpreters usually deliver more than 70\% of the necessary information~\cite{chmiel2021effects} and 95\% ideally. Thus in this paper, we use 80\% to indicate high-level human interpreters. 

Motivated by the huge success of LLMs in machine translation \cite{achiam2023gpt, brown2020gpt3} and speech translation \cite{chu2023qwen, huang2023speechtranslationlargelanguage, reid2024gemini}, we propose to employ LLMs to accomplish the SiST task.
Specifically, we identify three primary challenges.
First, a key challenge for incorporating LLM into the SiST is the read-write policy, where LLM needs to provide partial translation for input speech. 
Second, achieving human equivalent performance requires understanding and translation of terminologies and uncommon phrases that LLMs cannot learn from training data. 
Lastly, the scarcity of training data continues to hinder the performance on the SiST task.

\begin{figure}[t]
    \centering
    \includegraphics[width=1\linewidth]{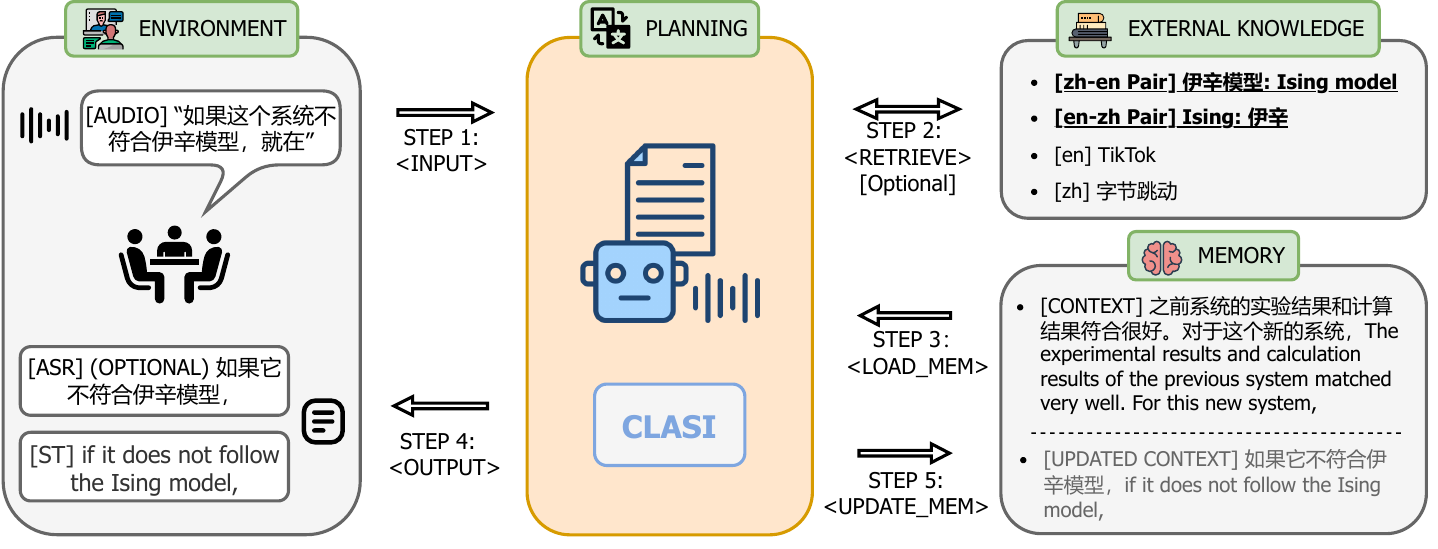}
    \caption{Overall framework of {\method}. The process begins in Step 1, where {\method} processes the incoming audio data. Optionally, the retriever is activated to obtain the relevant information from the external knowledge database. For instance, translating ``\underline{伊辛模型}'' to ``\underline{Ising model}'' for accurate speech translation. Step 3 involves accessing transcription~(optional) and translation in the last round memory. Steps 4 and 5 entail using the Chain-of-Thought (CoT) method to generate both the transcription~(optional) and translation, followed by a memory update. The cycle then repeats from Step 1 for the subsequent speech segment.}
    \label{fig:framework}
\end{figure}

To address these challenges, we introduce our end-to-end approach, {\method}, a \textbf{C}ross-\textbf{L}ingual \textbf{A}gent that accomplishes \textbf{S}imultaneous \textbf{I}nterpretation by iteratively performing multiple actions, as illustrated in Figure \ref{fig:framework}. 
Regarding the first challenge, we imitate professional human interpreters to learn their policy of segmenting a complete sentence into several semantic ``chunks'' through syntactic boundaries (pauses, commas, conjunctions, etc.) and contextual meaning. 
To enable \method to learn such a policy, we follow a data-driven policy learning process and invite human interpreters to annotate real-world speech, which includes the read-write timing for segmentation. From the data, {\method} learns the robust read-write policy for SiST from humans.

For the second challenge, we include two external modules to augment our {\method} agent: an external knowledge database that stores terminologies and paired translations, and a memory that stores the context of speech. 
However, the external knowledge database may contain tremendous terms that not only increase the inference time but may also lower the performance of our approach because of noisy intervention. Therefore, we propose a novel Multi-Modal Retrieval Augmented Generation (MM-RAG) process. 
A multi-modal retriever extracts knowledge from the external database based on the speech input. 
The retrieved information and the context from memory are then appended to the prompt of our LLM agent to augment the translation through in-context learning.

Addressing the data scarcity of the SiST task, we adopt a three-stage training methodology: pretraining, continual training, and fine-tuning. First, our LLM and audio encoder are independently pretrained on our large-size in-house datasets. 
Then, our model is continually trained with billions of tokens of mediocre-quality synthesized speech translation data, aiming to align the speech and text modalities.
We also include multiple tasks to enhance the in-context learning ability of LLM to better utilize the contextual information from the retriever and prior translation.
In the last stage, we fine-tune the model with a small amount of human-annotated data, further imitating professional human interpreters to improve the robustness and translation quality.

In addition, we would like to highlight that the conventional automatic evaluation metrics~\cite{papi2022over, papineni2002bleu, rei2020comet, sellam2020bleurt} of simultaneous interpretation might not be good indicators for reflecting the performance of SiST, which often contains compaction, abstraction, and paraphrasing. 
Aligned with human interpreters~\cite{moores2024nerle, wu2010assessing}, we propose a new evaluation metric named Valid Information Proportion (\humaneval)\footnote{Detailed guidelines of our proposed \humaneval metric can be found in Appendix~\ref{app:human_evaluation_guidelines}.}. VIP represents the percentage of information that can be precisely delivered, reflecting the central objective of SiST: communication in real-time.
Through thorough human evaluation on diverse and challenging real-world long speech datasets, our approach outperforms other currently accessible systems by a large margin.
As shown in Figure \ref{fig:fig1}, taking the Chinese-to-English direction as an example, {\method} achieves a \humaneval score of 81.3\%, significantly narrowing the gap between machine-assisted systems and human interpreters.

Our contributions can be summarized as follows:
\begin{itemize}
    \item We introduce our end-to-end approach, {\method}, an LLM agent that is designed to perform high-quality and human-like simultaneous translation. Through human evaluation, our approach demonstrates significantly better performance compared to existing accessible SiST systems.
    \item We propose a new data-driven read-write strategy by imitating professional human interpreters. Without the requirement of complicated human pre-design, the strategy could balance translation quality and latency effortlessly. Unlike most commercial systems where the outputs are frequently rewritten during the translation process for better quality, our strategy guarantees all the outputs are deterministic while maintaining high quality.
    \item Motivated by the preparatory trajectory of human interpreters, we introduce a novel Multi-Modal Retrieval Augmented Generation (MM-RAG) process that empowers the LLM with domain-specific knowledge in real time. The proposed module
    further improves the translation quality with minimal computational overhead during inference. 
    \item We work closely with professional human interpreters to develop our evaluation strategy, Valid Information Proportion (\humaneval), and detailed guidelines are open-sourced. Meanwhile, we release a human-annotated test set focusing on diverse real-world scenarios and long speech translations.
\end{itemize}

\section{Methods}
\begin{figure}[t]
    \centering
    \includegraphics[width=1\linewidth]{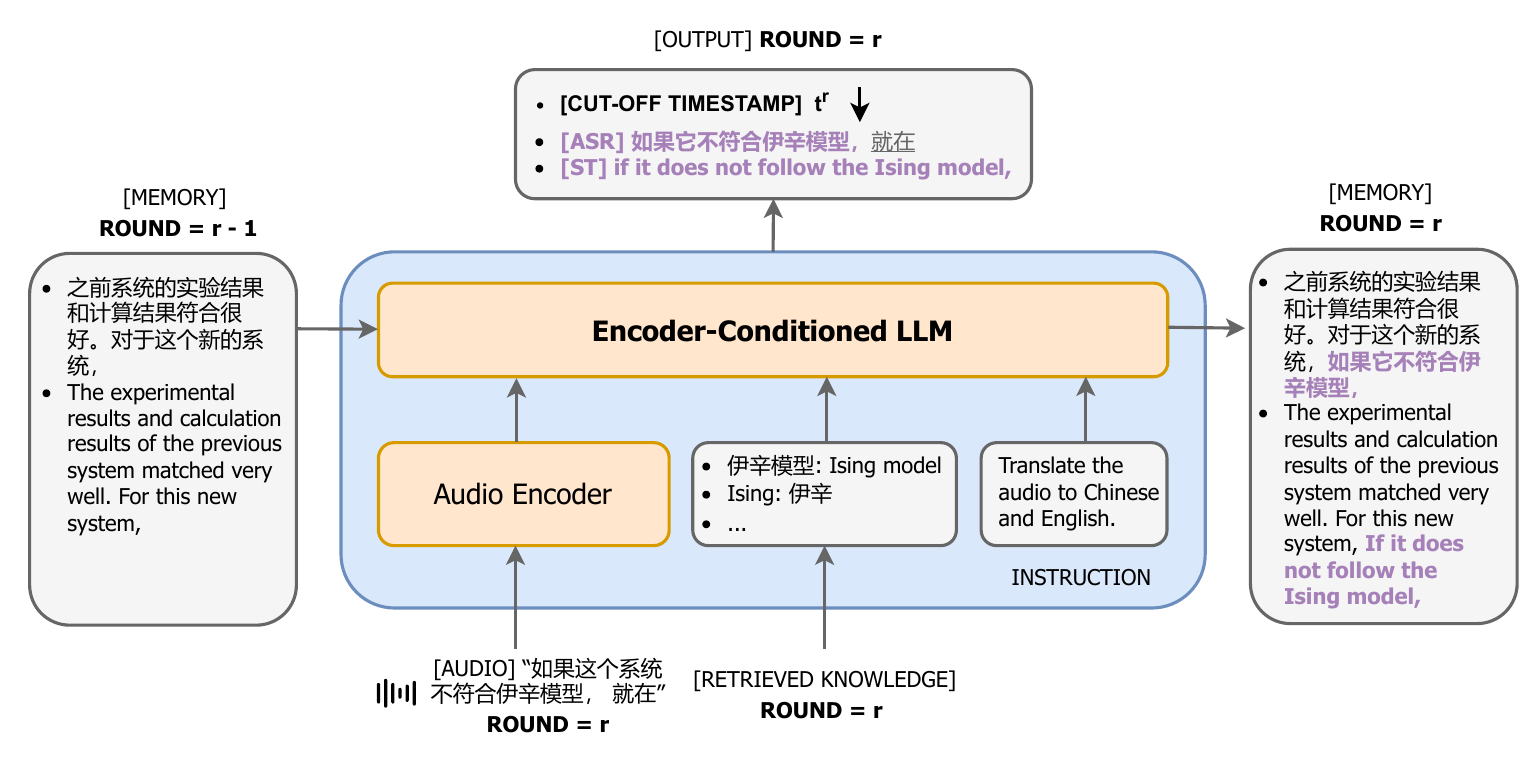}
    \caption{Architecture of \method agent. At round $r$, our model processes the current input audio stream alongside the memory from the previous round ($r-1$), and any retrieved knowledge. \method generates a response based on specified instructions and concurrently updates its memory. Additionally, the model determines the cut-off timestamp of the last semantic chunk. For instance, in the provided example, the phrase preceding ``\underline{就在}'' is identified as a complete semantic chunk, with the cut-off timestamp positioned right after this phrase. }
    \label{fig:framework_d}
\end{figure}

\subsection{Framework}
\Cref{fig:framework} presents a flow of operation of {\method}. To perform the SiST task, we design 5 operations: \texttt{<INPUT>}, \texttt{<OUTPUT>}, \texttt{<RETRIEVE>},  \texttt{<LOAD\_MEM>}, and \texttt{<UPDATE\_MEM>}. The following sections describe the details of each operation.
As further illustrated in \Cref{fig:framework_d}, {\method} is an LLM agent that can take input speech, instruction, relevant information retrieved from external knowledge, and last round memory as context. The memory stores previous transcriptions (optional) and translations. At round $r$, it first reads speech $\rvx_{t^{r-1}:T^r}$, where $t^{r-1}$ is the predicted cut-off time of round $r-1$ and $T^r$ is the end time for audio stream at round $r$. 
Then the agent retrieves relevant information $\rvk_r$ from the external knowledge and loads context $\rvy_{1:r-1}$ from the last round memory. 
Once \method ``think'' sufficient context is loaded, it generates the transcription (optional), translation, and cut-off timestamp $t^r$: 
\begin{equation}
    \rvy_{r}; t^r = \text{TextDecoder}(\rvx_{t^{r-1}:T^r}, \rvk_r, \rvy_{1:r-1})\;
\end{equation}
where $t^r$ is the predicted cut-off timestamp indicating the end time for the current translation round $r$. 
$\rvy_r$ is then forwarded to update the memory.
When instructed to output the transcription, the LLM optionally engages CoT to generate transcription first and then the speech translation.
For the following round $r+1$, the audio stream begins with the predicted cut-off timestamp $t^r$.
\subsection{Architecture}
{\method} employs an Encoder-Conditioned LLM architecture. As shown in \Cref{fig:framework_d}, the audio encoder transforms input speech stream $\rvx$ to a series of continuous representations $\rvs$. 
Then, the LLM takes the speech representation $\rvs$, retrieved knowledge $\rvk$, historical translation $\rvy$ and instruction $\rvI$ as a sequence of prompt $(\rvy, \rvs, \rvk, \rvI)$ to generate the translation result $\rvy$. 

\noindent\textbf{Audio Encoder.} The audio encoder module contains a large-scale speech conformer \citep{gulati2020conformer} pretrained on millions of hours of speech data to achieve human parity performance on ASR, and an audio adapter to connect the audio encoder and LLM. 
The adapter downsamples the speech representations and the resulting representations are linearly projected to match the dimension of the LLM embedding layer. 
The projected speech representations lower the computational latency for SiST.

\noindent\textbf{Large Language Model.} The language model\footnote{We use Doubao LLM as our foundation model.} is a medium size decoder-only transformer \cite{vaswani2017attention} to balance performance and computation efficiency. It is pretrained on a large amount of text data and fine-tuned with instructions. The LLM directly takes the continuous embedding from both the audio encoder and text embedder as input. It autoregressively generates the transcription and translation response of the provided speech stream. 

\noindent \textbf{Multi-Modal Retriever.} 
The multi-modal retriever framework employs audio and text encoders to independently encode the audio stream and text key of the terminologies in the external knowledge database.
To enhance the alignment between audio embeddings and text embeddings, we incorporate an embedding fusion layer, which includes a multi-head attention module followed by a pooling layer.
The resulting pooled representation is subsequently fed into a linear projection layer to produce the final scores, indicating the probability of the text key's presence in the audio stream.
Terminologies with top scores are forwarded to the {\method} agent to enhance the translation quality.

\subsection{Data Driven Read-Write Policy: \texttt{<INPUT>} and \texttt{<OUTPUT>}}

Unlike predetermined read-write probabilities and heuristic waiting policies detailed in prior research~\cite{barrault2023seamless, ma-etal-2019-stacl, li-etal-2023-hw-tsc}, interpreters engage in a dynamic process of listening (read) and translating (write).
They attentively listen to the speaker's speech and segment lengthy sentences into semantic chunks, representing the smallest linguistic units capable of conveying a complete thought independently \cite{jones2014conference}.
Upon identifying a chunk that encapsulates sufficient information, they proceed to translate this segment into the target language, thereby providing an accurate and contextually appropriate translation.

Emulating the strategies of human interpreters, {\method} does not require to explicitly define the read-write policy. {\method} imitates their policies by waiting for complete semantic chunks. Specifically, given partial speech, {\method} only generates the translation for the complete chunks of the input speech. The model is trained with segmented speech data to learn such ability.
Mathematically, given source audio $\rvx_{1:M}$, we segment its translation into a series of $n$ ``chunks'' $\rvy_{1:n}$ and obtain the corresponding pair $\{(\rvx_{t^j:t^{j+1}}, \rvy_j\}_{j=1}^n$, where $\rvx_{t^j:t^{j+1}}$ and $\rvy_j$ denote the $j$-th segment of the audio and the corresponding translation. For training, our objective is to output all complete segmented translations and the cut-off time given random partial input audio $\rvx_{1:t}$
\begin{equation}
    \min \mathbb{E}_{t \sim \mathcal{U}[1, M]} -\log p_\theta(\rvy_{1:j}; t^j|\rvx_{1:t}) \quad j = \max_j\{j|t^j < t\}\;
    \label{eqn:read-write-train}
\end{equation}
where $\mathcal{U}$ indicates uniform distribution over time of speech. 
Trained with \Cref{eqn:read-write-train}, \method learns to generate the cut-off time for the input speech. Additionally, the objective function makes the \method wait for appropriate time before starting translation as the LLM will output nothing when it ``think'' current speech stream does not contain a complete speech chunk.

\subsection{Context Information: \texttt{<LOAD\_MEM>} and \texttt{<UPDATE\_MEM>}}

The memory stores translations and transcriptions in previous rounds $\rvy_{1:r-1}$. It has two functions. Firstly, it works with the input speech to determine which part of the speech has been translated and which part has not, helping \method make the read-write decisions and outputs the translation of the unfinished parts. 
Secondly, understanding human speech often requires context. For example, when a speaker talks about ``barrel bridge'', it often refers to the bridges built upon rivers that are supported by barrels. However, in the context of ``watch'', it refers to a mechanical structure in the watch. The phenomenon of polysemy in different contexts can lead to vastly different translation outcomes. Therefore, {\method} should be able to retrieve the context of the long speech for translating some keywords, and make appropriate translations under different contexts. 

As shown in \Cref{fig:framework_d}, at round $r$, \texttt{<LOAD\_MEM>} forwards relevant translations $\rvy_{1:r-1}$ to the LLM as a prompt. After \method agent generates the translation $\rvy_r$, \texttt{<UPDATE\_MEM>} stores it to the memory and obtains $\rvy_{1:r}$.

\subsection{Multi-Modal Retrieval Augmented Generation: \texttt{<RETRIEVE>}}

In real-world scenarios, the accurate speech transcription or translation of professional and domain-specific terminologies is challenging. Even human interpreters require prior domain knowledge to understand those terminologies, including names of people, locations, jargon, or special in-domain terms.
For example, an interpreter unfamiliar with the machine learning theory may not recognize the word ``Rademacher complexity'' when hearing it. 
Therefore, in various scenarios, human interpreters often prepare in advance to get familiar with the corresponding domain knowledge.

Motivated by the preparatory trajectory of human interpreters, we propose to integrate an external database to empower LLM with necessary domain-specific knowledge. Each item in the database contains a key and the corresponding value in text modality. The key, which may appear in the speech, is used as the input for the retriever. The value of the item may be itself, a paired translation of the target language, or even an explanation of the key. 

Theoretically, all items in the external database might be added into the prompt to provide information for the translation. However, the external knowledge database often contains tremendous items.
Simply prompting LLM with all the terms not only increases the
inference time but may also hurt the performance of \method because of noisy intervention. 
Therefore, we design a novel Multi-Modal Retrieval Augmented Generation (MM-RAG) process.
Our multi-modal retriever first retrieves the relevant terminologies from the database based on the input speech.
A small number of filtered items are incorporated into the prompt of {\method} agent for in-context learning as shown in \Cref{fig:framework_d}. 

With the retrieved knowledge and previous context from the memory, our LLM has the in-context learning ability to better utilize the provided contextual information.
To achieve this, we collect a series of in-context learning data to train the model. Compared with the previous approaches for intervention, such as shallow-fusion \cite{Borgeaud2021ImprovingLM, Kolehmainen2024MultiModalRF} and traditional substitution-based methods~\cite{li2019neural,luong2015addressing,wang2017sogou}, which generates fixed translation for given translation pair. Our method achieves better results and generates more coherent text. For example, in some internet companies, ``大盘 == overall performance'', while in most cases, it should be ``stock market''. Our method can choose the correct translation given different context. Besides, our method can use monolingual text from both source and target language to help the translation.

\section{Multi-Stage Training}
Our {\method} follows a multi-stage training process: pretraining, multi-task continual training, and multi-task supervised fine-tuning.  
In the first stage, the LLM and audio encoder are separately pretrained with massive amounts of in-house speech and text data. 
Next, a large amount of speech-text paired data is used to align audio and text modalities, building the fundamental capability for cross-modal multitasking. 
In the final stage, {\method} agent is fine-tuned with a small amount of human-annotated data to imitate the translation behavior of professional human interpreters.
Our multi-stage training process enables high efficiency of learning with a small amount of human-labeled data.

\subsection{Pretraining}
The in-house LLM and audio encoder are independently pretrained on different modalities of data.
The LLM follows a decoder-only transformer architecture and is first pretrained on a massive amount of monolingual and bilingual text data with cross-entropy loss and then fine-tuned on instruction-following data. The LLM performs excellently on various downstream tasks, especially translation tasks.

The audio encoder also follows a classic pretrain-finetune paradigm \citep{bai2024,baevski2022data2vec, hsu2021hubert, zhang2023usm} with a massive amount of speech-related data. 
The pretraining stage provides a proficiently trained LLM and audio encoder, setting a solid foundation for the following stages.

\subsection{Multi-task Continual Training}
\noindent\textbf{Training Method.}
We follow the work of \citep{huang2023speechtranslationlargelanguage} for multi-task training. Specifically, for streaming and higher-quality translation, we mainly focus on three tasks for training {\method}: Automatic Speech Recognition (ASR), Speech Translation (ST), and Text Translation (MT). 
To align the modalities of the pretrained LLM and audio encoder, 
{\method} is continually trained on various tasks with a substantial volume of paired data. 
We further strengthen the in-context learning ability of our approach by incorporating translation in the memory and knowledge from external databases.
As a result, we expand the ST tasks to different configurations as shown in \Cref{tab:sft1_tasks}. An ST translation can either be streaming or offline, direct or COT, with or without context, which leads to 8 different tasks.

\begin{table*}[h]
    \centering
    \resizebox{1.0\textwidth}{!}{
        \begin{tabular}{l l}
            \toprule
            \textbf{Configuration}  & \textbf{Explanation}  \\
            \midrule
            Direct & Speech is directly translated into the target language.  \\
            COT & Speech is first transcribed to the source language and then translated to the target language. \\
            \midrule
            Streaming & Given \textbf{partial} speech, translate segments with complete semantics to the target language. \\
            Offline & Given \textbf{complete} speech, translate the whole content to the target language.\\
            \midrule
            w/o Context & No historical translations and external knowledge are provided. \\
            w/ Context & Historical translations or external knowledge are provided as context. \\
            \bottomrule 
        \end{tabular}
    }
    \caption{Illustration of different configurations of ST task in Multi-task Continual Training.}
    \label{tab:sft1_tasks}
\end{table*}

\noindent\textbf{Training Data Construction.}
Both ASR and MT tasks have been last for a while, and there is a relatively large amount of ASR and MT data.
The major challenge of developing an end-to-end SiST model is the data scarcity of simultaneous ST. 
To this end, we propose a synthetic data construction pipeline. With a strong LLM, we synthesize two types of speech translation data for continual training: offline ST data and context-aware segmented streaming ST data.

1. Offline ST data: We mainly rely on ASR data to construct the offline ST data. Given the ground-truth transcription of speech, we use in-house LLM to translate the source language to target languages.
To ensure the readability and conciseness of the target language, the LLM is prompted to conduct Inverse Text Normalization (ITN), filler word smoothing, etc.

2. Context-aware segmented streaming ST data: The streaming ST data consists of fine-grained audio-text alignments and translation pairs for segmented semantic chunks.
Compared to offline ST data, streaming ST data
is even more challenging to collect. 
We find that human interpreter often segments long speech into a few semantic chunks, each of which can be translated independently to ensure an effective and smooth translation.
Motivated by such findings, we leverage LLM to construct streaming ST data by imitating the chunking process.
Long speech data are used to construct the streaming ST data, as the additional history can provide better contextual information.
First, we prompt the LLM to break down the ASR transcription into multiple independent semantic chunks, which are then translated into the target language. 
Subsequently, we align the semantic chunks with the corresponding audio chunks, obtaining the streaming ST data. 
Such data enable our model to handle incomplete speech inputs and generate partial translation in coherent semantics.

To measure the quality of the synthetic data, we conduct human evaluations based on our proposed \humaneval metric. 
The synthetic data achieves a \humaneval score of 81\%, satisfying the minimal requirement for further training. 

\subsection{Multi-task Supervised Fine-tuning}
\noindent\textbf{Training Method.}
Even though {\method} possesses a good translation quality on the SiST tasks after the previous multi-task continual training stage, we further boost the performance by fine-tuning on human-annotated streaming ST data with diverse tasks listed in \Cref{tab:sft1_tasks}. 
Such high-quality data enables our model to better align with the segmentation methodologies of professional human interpreters.
Furthermore, this process enhances our model's robustness to speech disfluencies such as stuttering, ensuring smoother communication in real-world scenarios.

\noindent\textbf{Training Data Construction.}
The source of human-annotated streaming ST data originates from real-world scenarios that contain various speech characteristics, such as disfluencies, stuttering, code-mixing, and specialized terminologies.
Such features ensure the robustness of our model in diverse conditions.
We engage professional human interpreters to provide high-quality annotations for simultaneous segmentation and interpretation of the speech data. 
Additionally, terminologies are identified and translated within the context, further strengthening the context-aware capabilities of {\method}. 

\subsection{Multi-Modal Retriever Training}
The multi-modal retriever is independently trained with a substantial dataset of speech recognition data. 
During training, words are randomly selected from speech transcription to serve as the positive sample, indicating their appearance in the speech.
Negative words are selected from different sentences, indicating the speech does not mention these words.
We assign a label of $1$ to positive samples and $0$ to negative samples, aiming to minimize the Binary Cross Entropy (BCE) loss. This approach helps refine the model's ability to distinguish relevant from irrelevant information, enhancing its overall performance and accuracy.
We label the positive sample as $1$ and the negative sample as 0, minimizing the Binary Cross Entropy (BCE) loss.
To evaluate the effectiveness of our retriever, we build an in-house retrieve development set. Each sample in the development set includes a short audio chunk and the mentioned terms in the audio. 
Note that the term here is defined as special keywords, such as name, location, abbreviation, and domain-specific word.

\section{Experiments}

\subsection{Evaluation Benchmark}
Quite a few number of evaluation benchmarks have been proposed for SiST over the past years, including MuST-C~\cite{di2019must}, FLUERS~\cite{conneau2023fleurs}, CoVoST~\cite{wang-etal-2020-covost,wang2020covost}, BSTC~\cite{zhang-etal-2021-bstc}, and GigaST~\cite{ye23b_interspeech}, etc. 
However, although much effort has been spent to build these benchmarks, they still suffer some shortcomings when facing real-world SiST applications. 

First, these benchmarks often contain speeches that are either recorded by volunteers (\textit{e.g.} CoVoST and FLUERS) or collected from formally, clearly, and fluently talk and podcasts by well-prepared speakers~(\textit{e.g.}  MuST-C and GigaST). In real-world scenarios such as online meetings or social media videos, the characteristics of the speech might inevitably be informal, unclear, or disfluent. 
Second, these benchmarks provide a shortcut for evaluating the translation quality by giving the manually segmented sentences. Such a shortcut offers a gap between the current benchmark and real-world applications, where the models might need to take long speech and conduct segmentation~\cite{anastasopoulos2021findings} by themselves. Consequently, evaluations on manually segmented datasets are likely to overestimate the performances of a real-world SiST system.
These discrepancies result in the evaluation on these benchmarks are not reliable for practical SiST systems. 

As a preliminary attempt to address the shortcomings as mentioned earlier, we propose a new benchmark~\benchmark for Chinese-to-English (zh-en) and English-to-Chinese (en-zh). \benchmark is collected from diverse sources, and most speakers talk naturally and casually without careful preparation.
We choose 10 popular domains: technology, healthcare, education, finance, law, environment, entertainment, science, sports, and art.
One video clip is selected for each domain from a well-known online video platform for both zh-en and en-zh settings.\footnote{\benchmark is available at \url{https://github.com/byteresearchcla/RealSI}. We do not own the copyright of the videos and only release our annotations together with the publicly available website links of the corresponding videos. If anyone believes that the content constitutes infringement, please contact us. We will remove the relevant content as soon as it is confirmed.}
Each sample in \benchmark is a nearly 5-minute speech to mock SiST without manual segmentation.
For systems that cannot take long-form audio as input, we also provide sentence-level timestamps for segmentation. 
\Cref{tab:dataset} presents the detailed statistics of our \benchmark.

\begin{table*}[h]
\centering
\resizebox{0.63\textwidth}{!}{
\begin{tabular}{lcccc}
    \toprule
    \multirow{2}{*}{\textbf{Domain}}& \multicolumn{2}{c}{\bfseries zh-en}  & \multicolumn{2}{c}{\bfseries en-zh}  \\
    \cmidrule(lr){2-3} \cmidrule(lr){4-5}
     & Duration & \#Segments & Duration & \#Segments \\
    \midrule
    Technology      & 5:23  & 51 & 3:25  & 31 \\
    Healthcare      & 3:16  & 30 & 3:34  & 22 \\
    Education       & 4:56  & 48 & 5:00  & 41 \\
    Finance         & 5:22  & 29 & 5:01  & 40 \\
    Law             & 4:38  & 49 & 4:48  & 29 \\
    Environment     & 4:18  & 34 & 4:24  & 31 \\
    Entertainment   & 5:16  & 53 & 5:12  & 39 \\
    Science         & 4:47  & 37 & 5:11  & 35 \\
    Sports          & 5:22  & 33 & 3:25  & 58 \\
    Art             & 7:54  & 67 & 4:17  & 21 \\
    \midrule
    Total           & 51:12 & 431 & 44:17  & 347  \\
    \bottomrule
    \end{tabular}
}
\caption{Statistics of our proposed {\benchmark} benchmark.}
\label{tab:dataset}
\end{table*}

\subsection{Baselines}
We compare \method with the open-sourced SiST model, \emph{SeamlessStreaming} \cite{barrault2023seamless}. In addition, because of the limited number of available SiST models, we choose to compare {\method} with several commercial systems. We denote the commercial systems as \emph{Commercial} 1-4. It is worth noting that unlike {\method}, most of the commercial SiST systems will first generate a temporary translation as soon as possible, then rewrite the temporary translation with a potentially better translation after getting more context.
Notwithstanding, we evaluate the finalized translation of these systems in all our experiments. 
Although this re-writing strategy could improve translation quality, continually revising existing translations might affect the user experience, potentially leading to additional confusion. 
We would also like to highlight that human interpreters usually do 
 not employ such a rewriting strategy during translation. 
 During the entire evaluation, we employ a general external knowledge database that maintains the same for all the evaluation in this paper. It does not contain domain-specific external knowledge to form unfair comparisons. The improvement of external knowledge is independently reported in \Cref{sec:mm-rag-exp}.

\subsection{Translation Quality}

\begin{table*}[h]
    \centering
    \resizebox{0.92\textwidth}{!}{
    \begin{NiceTabular}{lccccccc}[colortbl-like]
    \CodeBefore
        \columncolor{blue!5}{8} 
    \Body
        \toprule
        \multirow{2}{*}{\large \textbf{zh-en}}& \multicolumn{2}{c}{\small BLEU} & \multicolumn{2}{c}{\small BLEURT} & \multicolumn{2}{c}{\small COMET}  & \multirow{2}{*}{ {VIP$^{\dag}$  (\%)}} \\
        \cmidrule(lr){2-3} \cmidrule(lr){4-5} \cmidrule(lr){6-7} 
        & doc & sent & doc & sent & doc & sent &  \\
        \midrule
        \emph{SeamlessStreaming} & 11.3 & 8.9 & 33.9 & 42.7 & 75.9 & 65.9  &  13.2  \\ 
        \emph{Commercial 1} & 15.0 & 10.8 & 30.6 & 43.9 & 72.4 & 67.1 &  10.4  \\ 
        \emph{Commercial 2} & 19.6 & 15.0 & 37.7 & 53.4  & 79.8 & 75.8 &  14.6  \\
        \emph{Commercial 3}  & 24.5 & 19.9 & 40.2 & 56.4 & 81.8 & 78.9 &  25.0 \\
        \emph{Commercial 4} & 25.2 & 21.4 & 40.8 & 59.3 & 82.9 & 80.8 &  35.4 \\
        \midrule
        \emph{\textbf{\method}} & \bfseries 32.6 & \bfseries 28.1 & \bfseries 44.4 & \bfseries 65.9 & \bfseries 84.6 & \bfseries 84.7  &  {\textbf{81.3}$^*$}  \\
        \bottomrule
    \end{NiceTabular}
    }
    \resizebox{0.92\textwidth}{!}{
    \begin{NiceTabular}{lccccccc}[colortbl-like]
    \CodeBefore
        \columncolor{blue!5}{8} 
    \Body
        \toprule
        \multirow{2}{*}{\large \textbf{en-zh}}& \multicolumn{2}{c}{\small BLEU} & \multicolumn{2}{c}{\small BLEURT} & \multicolumn{2}{c}{\small COMET}  & \multirow{2}{*}{ {VIP$^{\dag}$  (\%)}} \\
        \cmidrule(lr){2-3} \cmidrule(lr){4-5} \cmidrule(lr){6-7} 
        & doc & sent & doc & sent & doc & sent &  \\
        \midrule
        \emph{SeamlessStreaming} & 14.8 & 10.4 & 22.1 & 27.8 & 64.6  & 60.6 &  2.0  \\ 
        \emph{Commercial 1} &  25.6 & 20.4 & 40.2 & 38.0 & 70.3 & 70.9 &  12.8 \\ 
        \emph{Commercial 2} &  29.6 & 26.0  & 50.5 & 46.8 & 78.2 & 75.8  &  16.8 \\
        \emph{Commercial 3} & 31.6 & 28.7 & 51.2 & 52.8 & 81.0 & 79.9 &  29.5 \\
        \emph{Commercial 4} & 29.8 & 26.4 & 47.0 & 48.0 & 77.5 & 76.7 &  41.6 \\
        \midrule
        \emph{\textbf{\method}} & \bfseries 37.4 & \bfseries 32.8 & \bfseries 54.2 & \bfseries 61.3 & \bfseries 87.4 & \bfseries 85.6 &  {\textbf{78.0}$^*$} \\
        \bottomrule
    \end{NiceTabular}
    }
    \caption{Experiment results of translation quality. Automatic evaluations were calculated on both document level and sentence level. In document level evaluations, each translation of a nearly 5-minute audio was considered one instance. For sentence level, automatic scores are calculated on human-segmented translations. \humaneval refers to the human-evaluated Valid Information Proportion that reflects the translation quality of these systems. 
    $^{\dag}$~Due to the limitations in human evaluation capacity, the \humaneval scores are calculated on 4 randomly selected samples out of 10 in \benchmark across all systems for fair comparison, while automatic metrics are evaluated on 10 samples. 
    $^*$~Additionally, we evaluate the performance of \method on all 10 samples, achieving \humaneval scores of 78.0\% for zh-en and 74.9\% for en-zh.
    }
    \label{tab:translation_quality}
\end{table*}

\textbf{Evaluation Metrics.}
Automatic evaluation metrics such as BLEU~\cite{papineni2002bleu}, BLEURT~\cite{sellam2020bleurt}, and COMET~\cite{rei2020comet} are widely used for evaluating the translation quality \cite{barrault2023seamless, iwslt-2023-international, iwslt-2022-international}. 
However, they may not be able to fully reflect the quality of the translation, especially for paragraph-level translation of long speech. It is argued that the current evaluation metrics are not sufficient for ST and SiST tasks \cite{gaido2024speech}. The work of~\cite{machavcek2023mt, wein2024barriers} also highlighted that there might be a discrepancy between automatic evaluation metrics with human evaluation. 

Therefore, besides the automatic evaluation, we collaborated with senior professional human simultaneous interpreters to standardize the guidelines for a more realistic human evaluation. 
Our proposed human evaluation metric focuses on whether the output of the translation model can accurately convey the speaker's original intention for each semantic fragment.
This is also the key objective of human interpreters in real-time translation. Note that a single semantic fragment indicates a complete piece of source speech. Typically, a single semantic fragment is a complete sentence. 
Detailed definition can be found in \Cref{app:eval_process}.
The percentage of valid information fragments within a complete speech session is defined as \humaneval, which is consistent with real-world criteria for human simultaneous interpretation~\cite{wu2010assessing}.

\noindent\textbf{Quantitative Analysis.} 
As shown in \Cref{tab:translation_quality}, we compare {\method} with the baseline methods on \benchmark dataset. 
In terms of the reliable human evaluation metrics, \humaneval, {\method} achieves scores of 81.3\% and 78.0\% for zh-en and en-zh, respectively. While all other models' \humaneval scores are lower than 42\%. For more references, we use 3 widely-used automatic evaluation metrics: BLEU\footnote{We use SacreBLEU~\cite{post-2018-call} for all the BLEU calculations in this paper.}, BLEURT, COMET. Under the automatic evaluation metrics, {\method} also surpasses baselines by a large margin. The detailed human evaluation results of \method can be found in \Cref{app:example_of_human_eval}. 

High \humaneval marks {\method} a practical system that can help listeners understand real-time speech without professional human interpreters. Note that we only consider a system is better than others when the \humaneval is higher. For example, even though \emph{Commercial 1} achieves higher scores than \emph{SeamlessStreaming} on BLEU and COMET, we still consider \emph{SeamlessStreaming} is a better system for zh-en translation based on \humaneval.

\subsection{Latency}
\noindent\textbf{Evaluation Metrics.} Due to the differences of grammatical structures between languages, a delay in simultaneous interpretation is inevitable. In this paper, we adopt the widely-used Average Lagging~(AL)~\cite{ma-etal-2019-stacl}, Length Adaptive Average Lagging (LAAL)~\cite{papi2022over} for comparing the latency of different methods. 
To achieve a fair comparison with systems that rewrite the translation, we calculate the time of the definite translation of these systems. 
We also propose an additional metric, First Letter Appearance Lagging (FLAL), to reflect user experience on each system.
FLAL represents the time that each system outputs the first determined translation. 

\noindent\textbf{Quantitative Results.} \Cref{tab:latency} compares the latency of our model with various systems in terms of AL, LAAL, and our proposed FLAL on the \benchmark and CoVoST.
We find that the existing metrics AL and LAAL are not suitable latency measurements of paragraph-level SiST on \benchmark. 
When the results are significantly shorter or longer than the reference translation, AL and LAAL may be largely exaggerated, leading to unreliable high latency. 
In these scenarios, FLAL is a more reliable and stable metric for all the systems.

Besides the paragraph-level latency evaluation, we compare our approach with other systems on the sentence-level dataset CoVoST2 zh-en, where both AL and LAAL produce reasonable values and the results are shown on the right side of  \Cref{tab:latency}.
Since the commercial systems usually rewrite the translation, their latency is higher than the \method. 
Compared with the fastest approach \emph{SeamlessStreaming}, \method achieves comparable latency but much better translation quality.

\begin{table*}[h]
    \centering
    \resizebox{1.0\textwidth}{!}{
    \begin{tabular}{lccccccccc}
        \toprule
        \multirow{2}{*}{\textbf{Model}} &  
        \multicolumn{3}{c}{\textbf{\benchmark (zh-en)}} & \multicolumn{3}{c}{\textbf{\benchmark (en-zh)}} & \multicolumn{3}{c}{\textbf{CoVoST2 (zh-en)}} \\
        \cmidrule(lr){2-4} \cmidrule(lr){5-7} \cmidrule(lr){8-10}
        & AL & LAAL & FLAL & AL & LAAL & FLAL & AL & LAAL & FLAL \\
        \midrule
        \emph{SeamlessStreaming} & 3.50 & 42.31 &  2.65 & 3.06 & 16.02 & 2.24 &  2.26 &  2.46 &  4.03  \\
        \emph{Commercial 1} &  2.10 & 13.22 & 3.27 & 4.53 & 20.71 & 1.88 & 3.05 & 3.26 &  4.01 \\
        \emph{Commercial 2} & 2.92 &  4.30 & 5.90 & 1.05 & 8.02 & 12.42 & 2.65 & 2.88 & 3.82 \\
        \emph{Commercial 3} & 12.31 & 12.65 & 15.70 & 8.45 & 15.81 & 9.68 & 3.67 & 3.86 & 6.14 \\
        \emph{Commercial 4} & 26.59 & 27.17  & 6.62 & 16.94 & 24.47 & 5.73 & 3.53 & 3.71 & 6.20 \\
        \midrule
        \emph{\textbf{\method}} &  2.17 & 6.34 & 4.20 & 0.34 & 3.17 & 6.00 & 2.63 & 2.83 & 5.02 \\
        \bottomrule
    \end{tabular}}
    \caption{Comparison of latency between \method and baselines. AL and LAAL are standard metrics for measuring latency in sentence-level datasets. Even though AL and LAAL yield reliable results on the sentence-level CoVoST2 dataset, we argue that they are less effective for long speeches due to the complexity of long-speech translation.
    Therefore, we propose First Letter Appearance Lagging (FLAL), representing the time that each system outputs the first determined translation.}
    \label{tab:latency}
\end{table*}

\noindent\textbf{Discussion.} While existing works put a lot of emphasis on the latency-quality trade-off \cite{koshkin2024transllamallmbasedsimultaneoustranslation, papi2022attention}, human interpretation usually uses Ear-Voice-Span (EVS) to evaluate the lagging. EVS measures the average time from when the speaker finishes conveying a piece of information to when the audience hears the corresponding translation, which is similar to AL. The typical EVS of professional human interpreters usually ranges from 3 to 6 seconds~\cite{gumul2007time} to achieve high-quality translation. 

Consequently, we perform user studies and argue that the latency is less important than the translation quality for a practical SiST system. In the recent IWSLT 2023 simultaneous track \cite{iwslt-2023-international}, the ranking of models is also evaluated by the translation quality within certain latency constraints. We verify whether the latency of \method is acceptable to users through real-world user surveys. To the publication date of this paper, we collected 14 user surveys on zh-en direction, each user using \method for at least 30 minutes. Under the current latency performance shown in~\Cref{tab:latency}, only 1/14 == 7\% of them suggest that the latency significantly affects their user experiences while the rest think the improvement of translation quality outweighs the latency and overall output of \method largely helps them to understand the speech. Considering that the latency of \method is even lower than most of the commercial systems, We believe the latency of \method can be acceptable on most cases. 

Current latency metrics are proposed on sentence-level SiST. As shown in \Cref{tab:latency}, such metrics may not be suitable latency measurements for paragraph-level. 
As the importance of end-to-end evaluation for long speech keeps increasing, more refined metrics are required to measure the latency and provide a deeper insight into the systems.

\subsection{Supplementary Experiments}

To ensure a comprehensive evaluation of \method, our model is further evaluated on four additional datasets, including BSTC (zh-en)~\cite{zhang-etal-2021-bstc}, CoVoST2 (zh-en)~\cite{wang2020covost}, MuST-C (en-zh)~\cite{di2019must}, and GigaST (en-zh)~\cite{ye23b_interspeech}\footnote{We use the subset from in \href{https://github.com/SpeechTranslation/GigaS2S}{GigaS2S} for evaluation}.  
\Cref{tab:translation_quality_automatic_metrics_other} presents the results of the automatic evaluation metrics for both zh-en and en-zh. 
Due to the high cost of human evaluation, we are not able to provide VIP for these four datasets.
We observe that our model achieves consistently better performance than the baseline models.
Even though our system achieves the best automatic evaluation results among all the compared systems, 
we still would like to emphasize that such a sentence-level evaluation scheme might overestimate the performance of SiST systems.

\begin{table*}[h]
    \centering
    \resizebox{1.0\textwidth}{!}{
    \begin{tabular}{lcccccccccccc}
        \toprule
        \multirow{2}{*}{\textbf{Model}} &  \multicolumn{6}{c}{\textbf{BSTC zh-en}} & \multicolumn{6}{c}{\textbf{CoVoST2 zh-en}} \\
        \cmidrule(lr){2-7} \cmidrule(lr){8-13} &
         BLEU & BLEURT & COMET & AL & LAAL & FLAL &BLEU & BLEURT & COMET & AL & LAAL & FLAL \\
        \midrule
        \emph{SeamlessStreaming} & 9.7 & 34.4 & 78.2 & 11.41 & 68.92 & 3.50 & 19.3 & 54.7 & 77.1 &  2.27 &  2.46 & 4.03 \\
        \emph{Commercial 1} & 14.1 & 32.0 & 73.0  & 9.01 & 16.73 & 13.95 & 17.6 & 47.6 & 69.3 & 3.05 & 3.26 & 4.01 \\
        \emph{Commercial 2} & 17.6 & 39.2 & 81.2 & 6.35 &  7.92 & 13.04 & \bfseries 24.7 & 56.7 & 78.5 & 2.65 & 2.88 & 3.82 \\
        \emph{Commercial 3} & 21.5 & 41.6 & 83.7 & 12.88 & 13.63 & 22.55 & 24.2 & 54.1 & 75.9 & 3.67 & 3.86 & 6.14\\
        \emph{Commercial 4} & 21.2 & 41.9 & 82.3 & 30.50 & 31.84 & 9.61 & 22.1 & 56.1 & 76.8 & 3.53 & 3.71 & 6.20\\
        \cmidrule(lr){1-13}
        \emph{\textbf{\method}}& \bfseries 25.6 & \bfseries 44.8 & \bfseries 85.6 & 4.68 & 9.03 & 13.13 & 24.2 & \bfseries 56.8 & \bfseries 81.0 & 2.63 & 2.83 & 5.02 \\
        \midrule
        \midrule
        \multirow{2}{*}{\textbf{Model}} &  \multicolumn{6}{c}{\textbf{MuST-C en-zh}} & \multicolumn{6}{c}{\textbf{GigaST en-zh}} \\
        \cmidrule(lr){2-7} \cmidrule(lr){8-13} &
         BLEU & BLEURT & COMET & AL & LAAL & FLAL &BLEU & BLEURT & COMET & AL & LAAL & FLAL \\
        \midrule
        \emph{SeamlessStreaming} & 17.4 & 48.2 & 75.2 & 1.43 & 1.69 & 2.06 &  26.3 & 48.9  & 75.4 & 1.41 & 1.57 & 2.16 \\
        \emph{Commercial 1} & 24.0  & 55.2 & 81.2 & 2.62 & 2.91 & 2.07 & 43.1 & 59.6 & 83.4 & 2.55 & 2.73 & 2.33 \\
        \emph{Commercial 2} & \bfseries 28.2 & 59.5 & 83.1 & 3.25 & 3.51 & 4.84 & 45.7 & 63.2 & 85.0 & 3.13 & 3.28 & 5.12 \\
        \emph{Commercial 3} & 26.9 & 59.9 & 83.7 & 3.59 & 3.90 &  4.86 &  48.3 & 66.2 & 86.7 & 3.18 & 3.36 & 4.97 \\
        \emph{Commercial 4} & 27.3 & 60.0 & 83.4 & 3.25 & 3.54 & 4.86 & 43.3 & 59.9 & 83.5 & 3.06  & 3.23 & 5.00 \\
        \cmidrule(lr){1-13}
        \emph{\textbf{\method}} & 26.6 & \bfseries 61.8 & \bfseries 85.2 & 3.76 & 3.90 & 4.97 & \bfseries 50.4 & \bfseries 69.0 & \bfseries 88.8 & 3.30 & 3.40 & 5.01 \\
        \bottomrule
    \end{tabular}}
    \caption{Comparisons of \method and baselines on paragraph-level (BSTC) and sentence-level (CoVoST2, MuST-C, and GigaST) zh-en and en-zh datasets in terms of automatic evaluation metrics. We would like to emphasize that sentence-level evaluation schemes by automatic metrics cannot truly reflect the models' performance. \humaneval in \Cref{tab:translation_quality} is a better metrics for comparing different systems.}
    \label{tab:translation_quality_automatic_metrics_other}
\end{table*}

\subsection{MM-RAG Performance}
\label{sec:mm-rag-exp}
\subsubsection{Retriever}
Table \ref{tab:retrieve} presents the performance of various retrieve models on the development set of our proprietary dataset. Each sample in the test set includes a short audio chunk and the mentioned terms in the audio. 
Our MM-RAG retriever outperforms other open-source models by a large margin, achieving 91.3 \% vs. 26.0\% for Top-10 retrieve accuracy.
We compare two types of methodologies: audio-to-audio and audio-to-text. 
In the audio-to-audio approach, a Text-to-Speech (TTS) model is utilized to convert the text keys from the external knowledge database into audio format,
 forming a database with audio-based keys. 
The audio keys and the user-input audio are then encoded with the ASR model to produce the corresponding representations. 
The Top-$k$ retrieved items are subsequently determined using the Maximum Inner Product Search (MIPS) algorithm.
For audio-to-text approach, we compare MM-RAG with CLAP \cite{Elizalde2023CLAPLA}. As indicated in Table \ref{tab:retrieve}, the effectiveness of these models remains significantly below acceptable standards, and MM-RAG significantly outperforms them.

It is worth noting that the same audio encoder employed in our \method is utilized for generating audio embedding in the MM-RAG retriever.
Such a design ensures that the integration 
brings minimal computational latency to the overall framework.

\begin{table*}[h]
\centering
\resizebox{0.8\textwidth}{!}{
    \begin{tabular}{lccccc}
        \toprule
        \textbf{Model} & \textbf{Method} & \textbf{Finetuned} & \textbf{Top-1} &\textbf{Top-5} &\textbf{Top-10} \\
        \midrule
        \emph{CLAP} \cite{Elizalde2023CLAPLA} & Audio-to-Audio & No & 2.1 & 7.3 & 13.8 \\
        \emph{Wav2Clip} \cite{Wu2021Wav2CLIPLR} & Audio-to-Audio & No & 3.3 & 9.6 & 16.3 \\
        \emph{Whisper} \cite{Radford2022RobustSR} & Audio-to-Audio & No & 2.6 & 9.7 & 15.1 \\
        \emph{In-house ASR} & Audio-to-Audio & No & 7.2 & 19.4 & 26.0 \\ 
        \midrule
        \emph{CLAP} \cite{Elizalde2023CLAPLA} & Audio-to-Text & No & 2.7 & 6.4 & 10.8 \\       
        \emph{\textbf{MM-RAG (Ours)}} & Audio-to-Text & Yes & \textbf{63.2} & \textbf{88.4} & \textbf{91.3} \\   
        \bottomrule
    \end{tabular}
}
\caption{Top-$k$ retrieve accuracy (\%).}
\label{tab:retrieve}
\end{table*}

\subsubsection{ICL Performance}

By incorporating the retrieved terms from the external knowledge database as contextual information, our model's in-context learning ability significantly improves the performance of speech translation for in-domain terminologies. \Cref{tab:icl_keywords} compares our method with the widely-used shallow fusion for intervention in the generated conclusion. When calculating the Recall, we input 1 ground-truth keyword with 9 similar negative words as context. When calculating the false positive rate for precision, we input 10 similar negative words as context. ICL is able to achieve a high recall rate with good precision, obtaining the highest F1 while shallow fusion only gets a recall rate that is only half of ICL.

Additionally, we conduct an ablation study on our MM-RAG module within terminology-intensive scenarios incorporating the whole \texttt{<RETRIEVE>} pipeline.
The incorporation of the external knowledge database results in a significant increment in the \humaneval score by about 10\%, highlighting the effectiveness of our proposed MM-RAG.

\begin{table*}[t]
    \centering
    \centering
    \begin{tabular}{lccc}
        \toprule
         & \textbf{Recall (\%)} & \textbf{Precision (\%)} & \textbf{F1 (\%)}\\
         \midrule
        Shallow-Fusion & 40.8 &  94.2 & 56.9 \\
        ICL + Shallow Fusion &   79.2 &  73.4 &  76.2 \\
        ICL  &  79.2 & 86.3 & \bfseries 82.6 \\
        \bottomrule
    \end{tabular}
    \caption{Recall and Precision of ICL and shallow-fusion for the intervention of the keywords. When calculating the Recall, we input 1 ground-truth keyword with 9 similar negative words as context. When calculating the false positive rate for precision, we input 10 similar negative words as context. }
    \label{tab:icl_keywords}
\end{table*}

\subsection{Case Study} 
We present case studies to show the ability of \method in translating complicated speech for zh-en and en-zh in \Cref{tab:case_study_zhen} and \Cref{tab:case_study_enzh}. We choose one of the most-performed cascaded systems \product for comparison. The \product adopted a cascaded approach for SiST and it is shown in \Cref{tab:translation_quality} to be one of the best previous SiST systems. Detailed explanations are described in the tables. For zh-en direction, we present cases regarding robustness to recognition errors, reasoning ability, and trending words translation. For the en-zh direction, we present cases regarding native, expressive, and accurate terminology translations. More cases are shown in \Cref{tab:case_study_more}.

\begin{table}[htbp]
    \centering
    \setlength{\abovecaptionskip}{0.4cm}
    \resizebox{1.0\textwidth}{!}{
    \begin{tabular}{l|p{11cm}}
        \toprule
        \multicolumn{2}{c}{\textbf{CASE 1: Robustness to recognition errors} } \\
        \midrule
        Golden Transcription    &   欧文两罚命中，四分\underline{分差\textsuperscript{1}}，\underline{不到最后\textsuperscript{2}} \\ 
        \midrule
        \product ASR  &   欧文两罚命中，四分\underline{分叉\textsuperscript{1}}，\underline{不到最后\textsuperscript{2}} \\    \midrule
        \product Translation  &   Irving hit two free throws and \underline{split\textsuperscript{1}} the four-point spread \underline{to the end,\textsuperscript{2}} \\    \midrule
        \method ASR             &   欧文两罚命中，四分\underline{分叉\textsuperscript{1}}，\underline{不到最后\textsuperscript{2}} \\
        \midrule
        \method Translation     &   Kyrie makes both free throws, \underline{a four-point gap\textsuperscript{1}}, \\
        \midrule
        Explanation             &   The word \underline{分差\textsuperscript{1}} is mis-transcripted to \underline{分叉\textsuperscript{1}}, which actually means ``branch'' or ``split'' in English. \method still generates the correct translation. After only hearing \underline{不到最后\textsuperscript{2}}, \method decides not to translate immediately and leaves \underline{不到最后\textsuperscript{2}} to the next translation because of lacking context. While \product translates it incorrectly. \\
        \midrule
        \midrule

        \multicolumn{2}{c}{\textbf{CASE 2: Reasoning Ability for Translation}} \\
        \midrule
        Golden Transcription    &    \underline{绍兴二十年\textsuperscript{1}}担任右正言，弹劾胡寅   \\  \midrule
        \product ASR          &      \underline{绍兴二十年\textsuperscript{1}}担任佑正言，弹劾胡莹。    \\ \midrule
        \product Translation  &      \underline{Shaoxing twenty years\textsuperscript{1}} as YouZhengYan, impeach Hu Ying.   \\ \midrule
        \method ASR             &    \underline{绍兴二十年\textsuperscript{1}}担任右正言，弹劾胡寅       \\ \midrule
        \method Translation     &    \underline{In 1150, during the 20th year of Emperor Gaozong's Shaoxing era\textsuperscript{1}}, he served as the Right Censor and impeached Hu Ying  \\ \midrule
        Explanation             &    Literally, \underline{绍兴二十年\textsuperscript{1}} could be translated as ``Shaoxing 20th Year'', while \method could understand the actual year of \underline{绍兴二十年\textsuperscript{1}} is AD 1150, the 20th year under the reign of Emperor Gaozong. \\ 

        \midrule
        \midrule
        \multicolumn{2}{c}{\textbf{CASE 3：Trending words or slangs}} \\
        \midrule
        Golden Transcription    &   我们常说，你们也太\underline{卷\textsuperscript{1}}了吧，别卷了，还是\underline{躺平\textsuperscript{2}}舒服。   \\  \midrule
        \product ASR  &   我们常说，你们也太\underline{卷\textsuperscript{1}}了吧，别卷了，还是\underline{躺平\textsuperscript{2}}舒服。 \\  \midrule
        \product Translation  &   We often say that you are \underline{too curly\textsuperscript{1}}, don't curl up, or \underline{lie down\textsuperscript{2}} comfortably. \\  \midrule
        \method ASR             &    我们常说，你们也太\underline{卷\textsuperscript{1}}了吧，别卷了，还是\underline{躺平\textsuperscript{2}}舒服。  \\ \midrule
        \method translation     &   We often say, ``You are \underline{too competitive\textsuperscript{1}}. Stop it. It's more comfortable to \underline{lie flat\textsuperscript{2}}.''    \\ \midrule
        Explanation             &   Although in Chinese \underline{卷\textsuperscript{1}} could be translated to ``curly'' in some cases, it actually means ``involution'' in this context. \method translates it to ``competitive'', which is acceptable. \underline{Lie flat\textsuperscript{2}} is comparable to \underline{lie down\textsuperscript{2}} for translating \underline{躺平\textsuperscript{2}}, but the whole sentence is translated more naturally by \method.  \\ 
        
        \bottomrule
        
    \end{tabular}
    }
    \caption{Comparison between \method and \product for zh-en direction.}
    \label{tab:case_study_zhen}
\end{table}

\begin{table}[htbp]
    \setlength{\abovecaptionskip}{0.4cm}
    \setlength{\belowcaptionskip}{-0.4cm}
    \centering
    \small
    \begin{tabular}{l|p{11.5cm}}
        \toprule
        \multicolumn{2}{c}{\textbf{CASE 1: Native and Accurate Translation}} \\ \midrule
        Golden Transcription    &  You can't think of it on a \underline{case-by-case\textsuperscript{1}} basis. Either we all have rights or \underline{not have rights. Right\textsuperscript{2}}.         \\  \midrule
        \product ASR            &  You can't think of it on a  \underline{case-by-case\textsuperscript{1}} basis. Either we all have rights or \underline{nut have rights. Right\textsuperscript{2}}         \\  \midrule
        \product Translation    &  你不能\underline{根据具体情况\textsuperscript{1}}来考虑。要么我们都有权利，要么\underline{疯子都有权利\textsuperscript{2}}，对吧？ \\  \midrule
        \method ASR             &  You can't think of it on a \underline{case-by-case\textsuperscript{1}} basis. Either we all have rights or \underline{not have rights. Right\textsuperscript{2}}.         \\  \midrule
        \method Translation     &  你不能\underline{就事论事\textsuperscript{1}}地考虑这个问题。 要么我们都有权利，要么我们\underline{都没有权利\textsuperscript{2}}。    \\  \midrule
        Explanation             &  Although the \product translation of \underline{case-by-case\textsuperscript{1}} is correct, \method uses \underline{就事论事\textsuperscript{1}}, a well-known Chinese idiom, which is more native. Besides, \product ASR mis-transcripted \underline{not have rights\textsuperscript{2}} as \underline{nut have rights\textsuperscript{2}}, leading to a completely non-sense translation.        \\  
        \midrule
        \midrule
        
        \multicolumn{2}{c}{\textbf{CASE 2: Expressive Translation}} \\ \midrule
        Golden Transcription    &   She was sobbing in fear that this test in a foreign language has been put in front of her.         \\  \midrule
        \product ASR            &   She was sobbing in fear that this test in a foreign language has been put in front of her.        \\  \midrule
        \product Translation    &   她哭了，害怕这个外语的测试摆在她面前，\\  \midrule
        \method ASR             &   She was sobbing in fear that this test in a foreign language has been put in front of her.        \\  \midrule
        \method Translation     &   她因为害怕这门外语考试而哭泣，    \\  \midrule
        Explanation             &   Theoretically, the \product translation is correct literally. However, it's not expressive for native Chinese speakers. \method translation is expressive, conveying the same meaning of the source English sentence, which means ``She was sobbing because of fearing the foreign language test.''  \\

        \midrule
        \midrule

        \multicolumn{2}{c}{\textbf{CASE 3: Named Entity, Terminology Recognition and Translation}} \\ \midrule
        Golden Transcription    &   So let's let me put the \underline{COVID-19\textsuperscript{1}} for example, so now we we know that there are a lot of people are infected and they have um positive antibody tests         \\  \midrule
        \product ASR            &   So let's let me put the \underline{CUBA 19\textsuperscript{1}} for example, so now we we know that there are a lot of people are infected and they have um positive, \underline{anybody? tests,\textsuperscript{2}}        \\  \midrule
        \product Translation    &   所以让我以\underline{古巴19人\textsuperscript{1}}为例。所以现在我们知道有很多人被感染，他们是阳性的。\underline{有人吗？测试，\textsuperscript{2}} \\ 
        \midrule
        \method ASR             &   So let's let me put the \underline{COVID nineteen\textsuperscript{1}} for example, so now we we know that there are a lot of people are infected and they have um \underline{positive antibody tests.\textsuperscript{2}}        \\  \midrule
        \method Translation     &   以\underline{Covid-19\textsuperscript{1}}为例。 我们现在知道 有很多人被感染了， 并且他们的\underline{抗体检测呈阳性。\textsuperscript{2}}     \\  \midrule
        Explanation             &   \product cannot correctly recognize \underline{COVID-19\textsuperscript{1}} and \underline{antibody tests\textsuperscript{2}}, while \method successfully recognize and translate. \\
        \bottomrule
    \end{tabular}
    \caption{Comparison between \method and \product for en-zh direction.}
    \label{tab:case_study_enzh}
\end{table}

\section{Related Work}
\paragraph{Large language model.}

The encoder-decoder architecture \cite{polak-etal-2023-towards, ye23b_interspeech} has been widely explored in early speech translation research, but with the advent of large language models~\cite{achiam2023gpt}, there has been a growing interest in employing decoder-only architectures \cite{fu2023decoderonlyencoderdecoderinterpretinglanguage, seide2024speechreallmrealtime} for sequence-to-sequence problems. While recent efforts have emerged in utilizing large language models for machine translation \cite{li2024mtpatcherselectiveextendableknowledge, li2024eliciting, zheng2024finetuninglargelanguagemodels,zhu2024multilingual} and speech translation \cite{chu2023qwenaudioadvancinguniversalaudio, huang2023speechtranslationlargelanguage, tang2024salmonngenerichearingabilities}, the application of such models in simultaneous translation tasks remains limited. Although there has been early attempts to utilize LLM for simultaneous machine translation \cite{agostinelli2023simul,hu2024gentranslate, koshkin2024transllamallmbasedsimultaneoustranslation,zhang2024streamspeech}, to the best of our knowledge, no existing work has been found that explores the utilization of large language models for end-to-end simultaneous speech translation with such remarkable improvement.

Furthermore, LLMs have demonstrated impressive capabilities in tasks such as instruction following~\cite{achiam2023gpt,bai2023qwen,feng2024agilenovelframeworkllm, touvron2023llama}, reasoning \cite{paul2024refinerreasoningfeedbackintermediate, shinn2023reflexionlanguageagentsverbal}, and planning \cite{qiao2024autoactautomaticagentlearning, ruan2023tptulargelanguagemodelbased}. Recent research studies have leveraged prompt engineering to develop remarkable LLM agents that autonomously tackle complex tasks in diverse environments~\cite{wang2024survey}.  
In our work, we empower the LLM to perform sequential instructions to accomplish the simulation speech translation task.

\paragraph{Simultaneous Speech Translation.}
One of the important components of simultaneous speech translation is the segmentation strategy, which determines how the speech frames are fed to the models.
Different strategies could affect the latency and performance of the translation. 
According to \cite{liu2024recentadvancesendtoendsimultaneous}, segmentation strategies can be classified into fixed-length, word-based, and adaptive segmentation. Fixed-length strategies~\cite{nguyen2021empiricalstudyendtoendsimultaneous} divide the speech into equally-length segments, while word-based strategies~\cite{ma-etal-2020-simulmt} identify word boundaries within the speech. Adaptive segmentation \cite{dong-etal-2022-learning} detects boundaries for speech units. Among these categories, our method utilizes a fixed-length strategy.

Regarding the read/wait policy, the Wait-k method \cite{ma-etal-2019-stacl} and its variants \cite{nguyen2021empiricalstudyendtoendsimultaneous, zeng-etal-2021-realtrans} have been extensively studied in the context of text translation and speech translation. In comparison to these approaches, which explicitly learn the generation of read/write signals, another line of research focus on how to leverage offline translation models \cite{yan-etal-2023-cmus}.
When utilizing an offline translation model for simultaneous translation, it is important to address the stabilization of generated hypotheses to prevent excessive content refreshing experienced by the user.
\cite{liu2020lowlatencysequencetosequencespeechrecognition} first proposed a local agreement policy to stabilize the partial hypothesis, while \cite{Polk2023} introduced an incremental blockwise beam-search algorithm. In contrast to these methods, we enforce our model to generate consistent hypotheses by constraining the prompt to the language model.

For the model architecture, there are two primary methods for implementing speech translation systems: cascaded solutions \cite{guo-etal-2023-hw, iranzo2021streaming} and end-to-end solutions \cite{fukuda-etal-2023-naist, papi-etal-2023-direct, polak-etal-2023-towards}. Cascaded solutions involve separated ASR and MT components, while end-to-end solutions directly map speech to translations. Cascaded systems benefit from established techniques but suffer from latency and error propagation. End-to-end models offer real-time translation and improved quality through deep learning but require large-size training data. 
In our work, we implement an end-to-end model which combines the capabilities of ASR, MT, and ST.

\paragraph{Human Evaluation.} 
In the realm of speech translation, the choice of evaluation metrics plays a crucial role in assessing the quality and effectiveness of translation systems. While automatic metrics, such as BLEU~\cite{papineni2002bleu}, BLEURT~\cite{sellam2020bleurt}, and COMET~\cite{rei2020comet}, have traditionally been relied upon for evaluation, there is a growing recognition that they may not be the most suitable or comprehensive measure of performance \cite{marie-etal-2021-scientific}.
We observe that in more recent works \cite{barrault2023seamless, liu2024chatqasurpassinggpt4conversational, nllbteam2022languageleftbehindscaling, wang-etal-2023-document-level}, there is an increasing trend of evaluating systems using human assessments, particularly when LLM is employed in the work.
While human evaluation requires more resources and time compared to automatic metrics, its benefits outweigh the drawbacks. By incorporating human judgment, speech translation systems can be refined and optimized to align with user expectations, ensuring translations that are not only technically accurate but also linguistically and contextually appropriate.
In contrast to the existing human evaluation metrics, e.g.``continuous rating'' \cite{javorsky-etal-2022-continuous} and MQM (Multidimensional Quality Metrics) \cite{lommel2014multidimensional}, we have taken inspiration from professional human interpreters \cite{moores2024nerle} and propose to use VIP (Valid Information Proportion) as a human evaluation metric which precisely reflects the goal of the simultaneous translation task.


\section{Conclusion}
In this work, we introduce \textbf{C}ross \textbf{L}anguage \textbf{A}gent - \textbf{S}imultaneous \textbf{I}nterpretation, \method, an LLM agent to produce end-to-end simultaneous speech translation. Benefits from massive pretraining and imitation learning, \method achieves significantly better performances than state-of-the-art systems. Take the Chinese-to-English direction as an example, under strict and challenging human-evaluated metrics proposed by professional human interpreters, Valid Information Proportion (\humaneval), {\method} significantly outperforms baselines by a large margin. While all other systems obtain \humaneval by less than 40\%, \method achieves a VIP of 81.3\%, demonstrating human parity performance.
More specifically, we propose the following crucial components for the supreme performance of \method: (1) An encoder-conditioned LLM agent architecture that performs high-quality or even human-parity SiST process through simple actions. (2) Imitation learning from human interpreters for a natural read-write policy balances translation quality and latency in a data-driven manner, without complex human pre-designing. Under such policy, \method achieves a stable output scheme, where each output is deterministic, thus potentially better user experience than most commercial systems. (3) Motivated by the preparatory trajectory of human interpreters, \method could perform in-context learning from historical translations and external knowledge to provide sufficient information for translation. 
With the powerful translation ability of \method,  we believe it can further make cross-lingual communication seamless across different places all over the world.

\section*{Limitation and Future Work}
Although we achieved significant improvements over commercial systems in Chinese-to-English and English-to-Chinese tasks, more languages should be considered in the future. 
In our current implementation, \method performs a full action sequence for each translation round. Some of the actions, e.g., \texttt{<RETRIEVE>} is optional for easy translation scenarios since the model is capable of translating correctly without the help of external knowledge. Training the model to better determine whether to skip unnecessary actions is a future direction. 
For a product-level system, even though the latency of \method is acceptable in most cases, how to reduce the translation latency without lowering the translation quality is still interesting and potentially helpful for user experience. 
Furthermore, we argue that the current automatic metrics are not comprehensive for SiST evaluation. Most of the quality measurements do not consider key information, which is crucial in SiST scenarios. As such, we proposed \humaneval for better human evaluation. Consequently, more reliable automatic quality and latency metrics should be proposed in the future as well. 
Reinforcement learning from human feedback (RLHF) has been proven to be effective in enhancing LLM performance. Although \method achieves significantly superior results than previous state-of-the-art systems, further studies on how to build better multi-modal reward models and better RL methods for SiST is also an important direction. Incorporating more modalities, for example, end-to-end speech-to-speech generation, or even end-to-end video-to-video generation are also promising research topics.

\section*{Social Impact}
The powerful SiST system \method can be applied to various scenarios to facilitate cross-lingual communications. For example, it can be deployed to various conferences or daily meetings to help listeners understand speech in different languages. It can also be deployed as a system-level translation module to help users watch videos that are conveyed in different languages. For online gaming, it can also help to bridge the gap of cross-lingual communication and connect people speaking different languages. A powerful SiST system with human parity performance may significantly improve the efficiency of professional human interpreters.

Despite the huge positive social impact that \method may bring, every coin has two sides. Neglecting some low-resource languages may also bring unfairness to some minorities. Resolving these problems needs further cooperation from the society. We leave more languages supporting as our future work.

\section*{Authorship and Acknowledgements}
All contributors are listed in alphabetical order by last name. Corresponding to this work can be sent to any core authors' email.

\textbf{Core Authors.} 
All core authors contributed equally to this work. 
\begin{multicols}{2}
\begin{itemize}
    \item Shanbo Cheng 
    \item Zhichao Huang
    \item Tom Ko
    \item Hang Li
    \item Ningxin Peng
    \item Lu Xu
    \item Qini Zhang
    \columnbreak
    \begin{itemize}[label={}]
        \setlength{\itemsep}{2.5pt}
        \item chengshanbo@bytedance.com
        \item zhichao.huang@bytedance.com
        \item tom.ko@bytedance.com
        \item lihang.lh@bytedance.com
        \item pengningxin@bytedance.com
        \item xu.lu1@bytedance.com
        \item qini.z@bytedance.com
    \end{itemize}
\end{itemize}
\end{multicols}

\textbf{Labeling, Evaluation and Interpretation Team.} 
Our data labeling and human evaluation team led by Yifu Li, made diligent efforts in all kinds of help needed, which is irreplaceable in the success of this project. Special thanks to the human interpreter team led by Anna Liu for providing insightful and comprehensive analysis on data labeling, human evaluation, and other recommendations. 
\begin{itemize}
    \item Jingwen Chen
    \item Xiaoya Chen
    \item Yifu Li
    \item Huiying Lin
    \item Anna Liu
\end{itemize}

\textbf{Engineering Team.} 
We collaborated with our engineering team led by Tingshuai Yan, their infrastructure support is crucial for this project.
\begin{itemize}
    \item Weicheng Fu
    \item Tingshuai Yan
    \item Liehao Zou
\end{itemize}

\textbf{Acknowledgements.} We appreciate the Speech Understanding team for all kinds of help, especially data sharing, thanks to their tremendous work for all the in-house data. We would like to express our deepest thanks to all the contributors to this project, their brilliant work guarantees the success of this project. We also want to thank Wenda Xu and Xi Xu for their suggestions on automatic evaluations.

\newpage
\medskip

\bibliographystyle{plain}

\bibliography{ref}

\appendix
\newpage

\section{Human Evaluation Guidelines}
\label{app:human_evaluation_guidelines}
In this Appendix, we provide detailed human evaluation guidelines which are formulated by professional human interpreters.
\subsection{Key Indicator} 
We define the key indicator as the “Valid Information Proportion,” denoted as \humaneval. This metric measures the proportion of valid semantic fragments within a complete speech session. A semantic fragment is deemed valid if it effectively conveys the core information, accurately representing the speaker’s original intent. Typically, one complete sentence is considered a single semantic fragment. \humaneval assesses the model’s ability to capture and communicate the essence of the spoken content. 

\subsection{Evaluation Process}
\label{app:eval_process}
First, the human evaluators segment the long translation result into semantic fragments according to the formal rules as follows:
\begin{itemize}
    \item \textbf{Semantic Completeness:} Each fragment should contain one complete concept or information point. For example, in a conference translation, an ideal semantic fragment often corresponds to a full sentence.
    \item \textbf{Natural Language Pauses:} Pauses that naturally occur in speech often indicate the boundaries of semantic fragments. During segmentation, natural pauses should be extensively considered and avoid irrational interruptions. Also, punctuation and conjunction in the text should be considered to maintain integrity and clarity of information.
    \item \textbf{Logical Coherence:} Each segment should contain information that are logically coherent and continuous. Conditional sentences, causative sentences, or both parts of antithesis sentences should be kept within the same segment.
    \item \textbf{Grammatical Completeness:} Each segment should include all necessary grammatical components (e.g., subject, verb, object) and have a complete grammatical structure.
    \item \textbf{Proper Information Density:} Each segment should have a moderate amount of information, avoiding information overload. It is recommended that each segment not exceed 50 words.
\end{itemize}

After segmentation, the human evaluators follow the instructions below to evaluate the validity of the semantic fragments:
\begin{itemize}
    \item \textbf{Key Information Recognition}. Key information refers to the content that can constitute core information, including but not limited to proper nouns, keywords, terminologies, sentence structures, etc.
    \item \textbf{Correctness Assessment}. Evaluators assess whether the translation of key information is accurate and successful in conveying the correct spoken intentions. Misinterpretations of the speaker’s words, inaccuracies in analyzing the context, or erroneous translations of specific terms can all contribute to the failure of the assessment. 
    \item \textbf{Expressiveness Assessment}. Evaluators assess whether the whole segment is translated accurately, comprehensibly, and expressively to humans. Assessing for any vague, ambiguous, or misleading statements. This indicator primarily evaluates the clarity, fluency, and intuitiveness of the translation, rather than its accuracy. Typically, verbosity, complex sentence structures, or challenging grammatical constructions that are unnecessary would reduce the expressiveness of the translation, thus leading to failure of the assessment. 
\end{itemize}
If the translation fails any of the above assessments, the translation will be marked as invalid. After the evaluators assessed all semantic fragments, the \humaneval could be simply calculated as dividing the number of valid semantic fragments by the total number of fragments.

We illustrate the evaluation criteria with two examples in Table~\ref{tab:human_eval_examples}. Although these translations achieve “high accuracy” in automatic evaluations, we still categorize them as invalid. It’s important to note that our standard aims to emulate human interpreters, presenting a significant challenge to both human evaluators and translation systems.

\begin{CJK*}{UTF8}{gbsn}
\begin{table*}[htbp]
    \setlength{\belowcaptionskip}{-0.4cm}
    \centering
    \small
    \begin{tabular}{l|p{11cm}}
        \toprule
        \multicolumn{2}{c}{\textbf{Example 1: Correctness Assessment} } \\
        \midrule
        Golden Chinese  &  请确保服务器后端的API接口完全遵循\underline{RESTful\textsuperscript{1}}架构原则。 \\    \midrule
        Reference       &  Please ensure the server backend's API interface fully complies with \underline{RESTful\textsuperscript{1}} architectural principles.  \\   \midrule
        Translation &  Please ensure the server backend's API interface fully complies with \underline{restless\textsuperscript{1}} architectural principles.  \\    \midrule
        Explanation     & Although the sentence-level BLEU score is near 80, the translation is still considered invalid, because the keyword "RESTful" is mistranslated.  \\   \midrule
        \midrule
        \multicolumn{2}{c}{\textbf{Example 2: Expressiveness Assessment} } \\
        \midrule
        Golden Chinese  &  这部分跟资源那边，前端资源这一块搞定了吗？ \\    \midrule
        Reference       &  Did you arrange the front-end resources well?  \\   \midrule
        Translation 1  &     Is this part related to the front-end resources? Did you finish the front-end resources?   \\    \midrule
        Translation 2    &   Has the front-end resource been settled?    \\ \midrule
        Explanation     &    Translation 1 is redundant, disfluent and contains minor errors, thus not easy to understand by human evaluators, while Translation 2 generated by \method is concise and fluent, and conveys the speaker's intention appropriately. \\
        \bottomrule
        
    \end{tabular}
    \caption{Human evaluation examples of Chinese-to-English Translation task.}
    \label{tab:human_eval_examples}
\end{table*}
\end{CJK*}

\subsection{Correlation with Automatic Metrics}
\begin{figure}[hbtp]
    \centering
    \begin{subfigure}[b]{0.49\textwidth}
        \includegraphics[width=\textwidth]{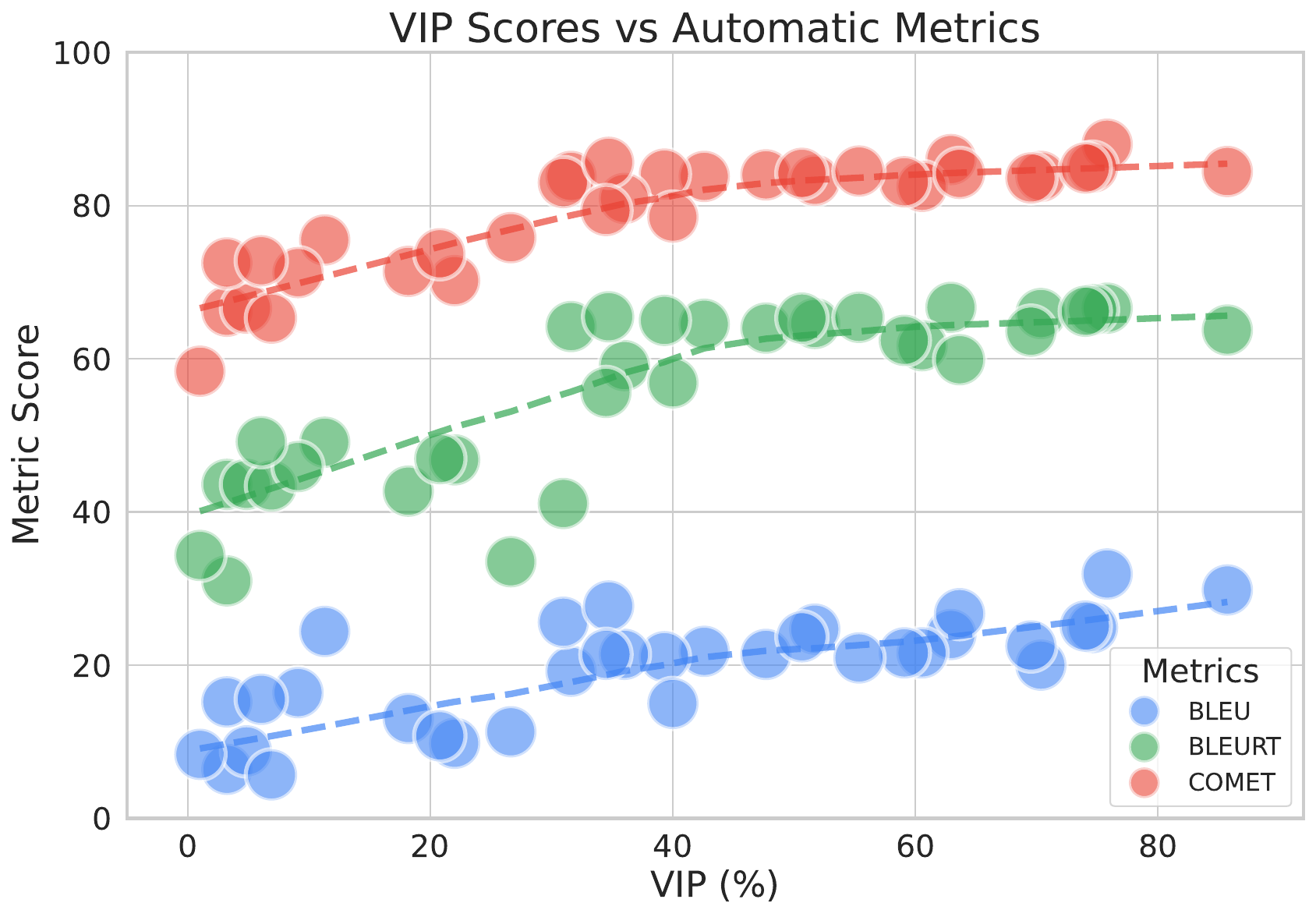}
        \label{fig:corr_scatter}
    \end{subfigure}
    \begin{subfigure}[b]{0.49\textwidth}
        \includegraphics[width=\textwidth]{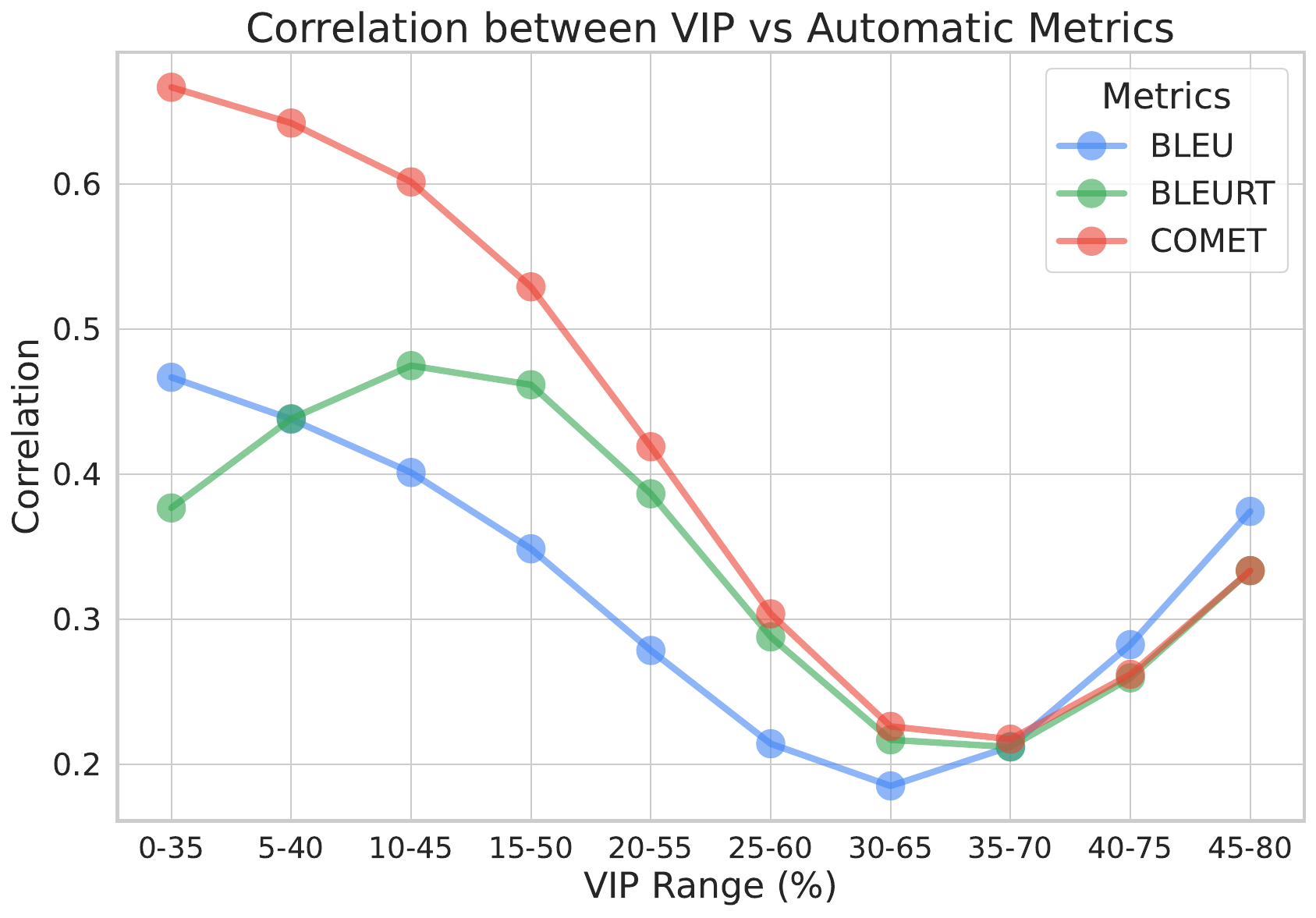}
        \label{fig:corr_curve}
    \end{subfigure}
    \caption{Analysis of \humaneval vs different automatic metrics on the zh-en direction. The distribution and regression curve of the data points for each metric are shown in the above-left figure. Line charts for the calculated correlation between \humaneval and Automatic metric within multiple intervals are shown in the right figure. Due to the limitation of human labeling capacity, we collect 35 rounds of human evaluation results for zh-en direction on our in-house testset. }
    \label{fig:humanvsauto}
\end{figure}

\Cref{fig:humanvsauto} shows the distribution and regression curve for \humaneval with regard to BLEU, BLEURT, and COMET, respectively. From the scatter points in \Cref{fig:corr_scatter} we may observe that as \humaneval score increases, the growth of the automatic metric curves slows down and becomes less significant, making it hard to reflect the real changes in translation quality. The correlation curves in~\Cref{fig:corr_curve} also demonstrate the finding. Here, we calculate Kendall's Tau correlation coefficient \cite{kendalltau}, which measures the monotonic correlation between two ordered variables. In low \humaneval ranges, the correlation between \humaneval and automatic metrics is observable; as the score increases, the correlation harshly drops, which indicates a significant distortion of the automatic metrics. A possible reason is that the translations may differ from the groundtruths by only a few words in the mediocre ranges. However, in a real simultaneous interpretation scenario, these words are likely to be keywords that play important roles in conveying precise information, which may significantly impact \humaneval scores.

\newpage
\section{Supplementary Materials}

\subsection{Supplementary Case Study}
We provide more case studies in \Cref{tab:case_study_more}. In terms of informal, disfluent, code-mixing, and named-entity translation, \method could achieve much better results than the commercial products. Benefits from the end-to-end approach, \method could also understand the original speech tone and generate better translations.

\begin{table}[h]
    \setlength{\abovecaptionskip}{0.4cm}
    \setlength{\belowcaptionskip}{-0.4cm}
    \centering
    \small
    \begin{tabular}{p{3cm}|p{12cm}}
        \toprule
        \multicolumn{2}{c}{\textbf{CASE 1： Informal, disfluent speech translation} } \\
        \midrule
        Golden Transcription    &   那基于这些观察，\underline{那我们是不是啊，我，我，我\textsuperscript{1}}，我们是不是可以去找一种，就是像GPT3.5一样的，统一的建模方法？ \\ \midrule
        \product ASR          &   那基于这些观察，\underline{那我们是不是啊？我们是不是我？\textsuperscript{1}}我们是不是可以去找一种，就是像\underline{gvt3.5\textsuperscript{2}}一样的，统一的建模方法？ \\ \midrule
        \product Translation  &   So based on these observations, \underline{are we? Are we me? Can we go\textsuperscript{1}}, like \underline{gvt3.5\textsuperscript{2}}, and do this unified modeling \\ \midrule
    
        \method ASR             &   那基于这些观察，\underline{那我们是不是啊，我，我，我，\textsuperscript{1}}我们是不是可以去找一种，就是像GPT3.5一样的，统一的建模方法？ \\ \midrule
        \method Translation     &   Based on these observations, \underline{can we find\textsuperscript{1}} a unified modeling method \underline{like GPT3.5\textsuperscript{2}}? \\
        \midrule 
        Explanation             &   The first labeled Chinese phrase (superscript 1) actually means \underline{"can we find"\textsuperscript{1}}. \method can generate a much more fluent, concise translation than the \product. Besides, for keyword \underline{GPT3.5\textsuperscript{2}}, \method can generate correct ASR and translation. \\ 
        \midrule
        \midrule
        
        \multicolumn{2}{c}{\textbf{CASE 2：Disfluent and code-mixing speech }} \\
        \midrule
        Golden Transcription    &   我听过一句话叫，\underline{pri, pri, prioritization, prioritization\textsuperscript{1}} is only real when it hurts。   \\  \midrule
        \product ASR  &   我听过一句话叫，\underline{Pro, 不管, prioritization, prioritization\textsuperscript{1}} is only real when it hurts.   \\  \midrule
        \product Translation  &   I heard a saying called \underline{Pro, Anyway, it is
Prioritization\textsuperscript{1}} is only real when it hurts.   \\  \midrule
        \method ASR             &   我听过一句话叫，\underline{priortizaiton, prioritization\textsuperscript{1}} is only real when it hurts。 \\
        \midrule
        \method Translation     &   I heard a saying that \underline{prioritization\textsuperscript{1}} is only real when it hurts. \\ \midrule
        Explanation             &   The speaker stutters when saying the English sentence, which is very common in real-world scenarios. \method can fully understand and generate the correct English text without any repetition. \\ 
        
        \midrule
        \midrule
        \multicolumn{2}{c}{\textbf{CASE 3: Named-entity recognition and translation}} \\
        \midrule
        Golden Transcription    &    好球！\underline{迪亚斯\textsuperscript{1}}的传中，C罗来争抢，这个就是\underline{C罗\textsuperscript{2}}最喜欢的   \\  \midrule
        \product ASR          &    好球,\underline{比亚斯\textsuperscript{1}}的传统，C罗来争抢这个就这是\underline{C罗\textsuperscript{2}}最喜欢的     \\ \midrule
        \product Translation  &    Nice shot, \underline{Bias'\textsuperscript{1}} traditional C Ronaldo to compete for this is \underline{C Luo's\textsuperscript{2}} favorite  \\ \midrule
        \method ASR             &    好球！\underline{迪亚斯\textsuperscript{1}}的传中，C罗来争抢。 这个就这是C罗最喜欢的  \\ \midrule
        \method Translation     &    Nice cross by \underline{Dias\textsuperscript{1}}, Ronaldo goes for it. This is \underline{Ronaldo's\textsuperscript{2}} favorite.   \\ \midrule
        Explanation             &    \product cannot correctly recognize the name of the famous football player, Ruben Dias. As for the name of Cristiano Ronaldo, although it translates correctly the first time, but fails the second time. \method can perform perfect recognition and translation. \\ 
        \midrule
        \midrule
        
        \multicolumn{2}{c}{\textbf{CASE 4: Speech tone understanding}} \\
        \midrule
        Golden Transcription    &    门前斜传，漂亮！这下漂亮！球进了！摆乌龙！哎，自摆乌龙。   \\  \midrule
        \product ASR          &    球没有斜转漂亮，这下球进了白骨龙，这白骨龙     \\ \midrule
        \product Translation  &    The ball didn't spin beautifully, and now the ball went into the bone dragon, the bone dragon  \\ \midrule
        \method ASR             &    门前斜转，漂亮！这下漂亮！球进了！摆乌龙！哎，自摆乌龙。  \\ \midrule
        \method Translation     &    Diagonal shot in front of the goal. Beautiful! What a beauty! Goal! Own goal! Yes, own goal.    \\ \midrule
        Explanation             &    \method could recognize the speaker with an exciting tone, thus generating the translation with exclamation marks. Besides, in this case, which is a complicate scenario of a football game, the ASR outputs of the \product are mostly incorrect, leads to nonsense translation.  \\ 
        \bottomrule
        
    \end{tabular}
    \caption{Comparision between \method and \product for zh-en direction.}
    \label{tab:case_study_more}
\end{table}

\subsection{Example of Detailed Evaluation Result on \benchmark}
\label{app:example_of_human_eval}
\begin{figure}[h]
    \centering
    \resizebox{1.0\textwidth}{!}{
        \includegraphics{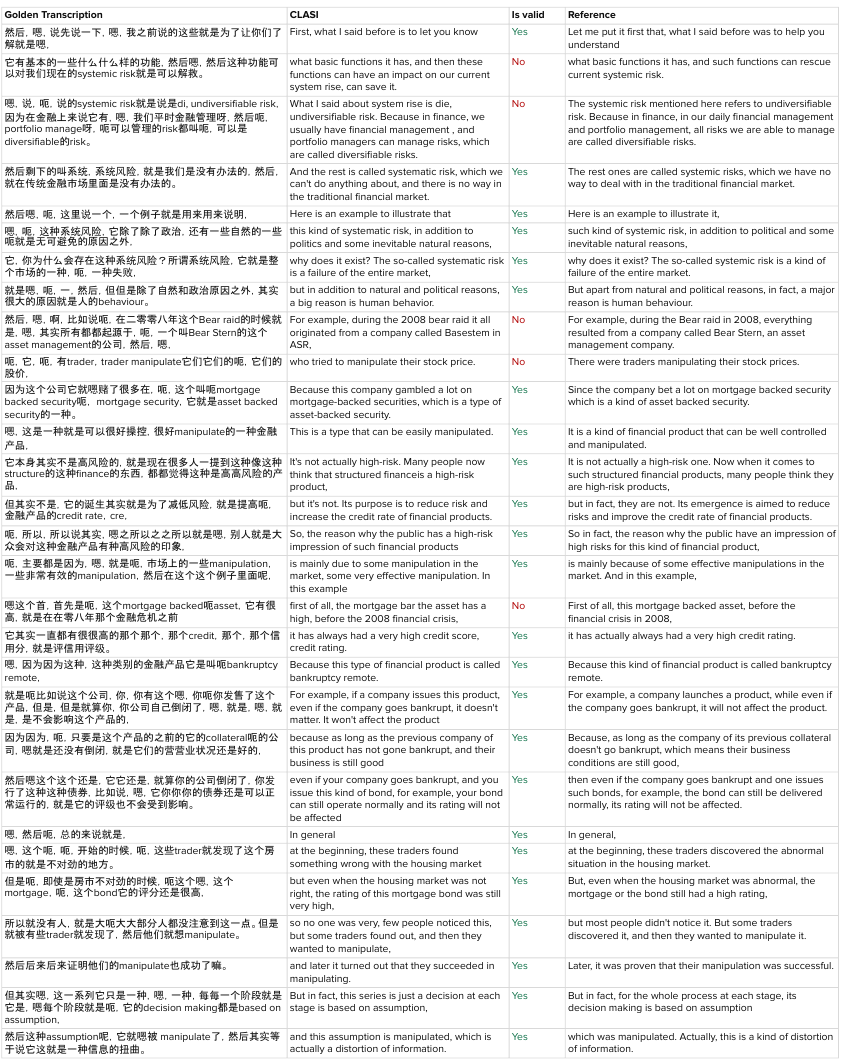}
    }
    \caption{The first column indicates the golden transcription of the source text. Each row indicates one semantic fragment split by human evaluators. The second column is the translation results of \method. The third and fourth columns indicate the validity of translation and reference translation, respectively. In this case, the \humaneval is 24/29 == 82.8\%.}
    \label{fig:human_example}
\end{figure}

We provide a detailed human evaluation result for \method in Figure~\ref{fig:human_example}, where we provide golden source transcription, \method output, human evaluation results, and reference translation. We randomly choose one of the test samples in \benchmark. We share the full detailed evaluation results at online sheets\footnote{We provide the full evaluation results of \method at \url{https://bit.ly/clasi-eval}}
for academic reference. Note that to ensure fair comparison, when evaluating multiple systems, we randomly shuffle the ordering between systems for each semantic fragment so that human evaluators cannot identify the specific system.

\end{CJK}

\end{document}